\documentclass[conference]{IEEEtran}
\IEEEoverridecommandlockouts
\def\blind{0}  % 0 or 1

\usepackage{cite}
\usepackage{amsmath,amssymb,amsfonts}
\usepackage{algorithmic}
\usepackage{graphicx}
\usepackage{adjustbox}
\usepackage{booktabs,multirow,makecell,array}
\usepackage{textcomp}
\usepackage{xcolor}
\usepackage[hidelinks,breaklinks]{hyperref}
\usepackage{cleveref}
\crefname{table}{Table}{Tables}
\Crefname{table}{Table}{Tables}
\crefname{figure}{Fig.}{Figs.}
\Crefname{figure}{Fig.}{Figs.}
\crefname{section}{Section}{Sections}
\Crefname{section}{Section}{Sections}
\usepackage{subcaption}
\usepackage{placeins}  % \FloatBarrier: keep all tables within the main content, before the references
\usepackage{tikz}
\usetikzlibrary{positioning,arrows.meta,calc}
\usepackage{orcidlink}

\graphicspath{{figures/}}

\newcommand{\blank}{\epsilon}
\newcommand{\R}{\mathbb{R}}

\begin{document}

% {*}Note: Sub-titles are not captured for https://ieeexplore.ieee.org  and should not be used
\title{Gradient-Based Speech-to-Text Alignment for Any ASR Model: From CTC to Speech LLMs
}

\if\blind1

% Anonymous review placeholder - remove upon acceptance
\author{
  \IEEEauthorblockN{Anonymous Authors}
  \IEEEauthorblockA{
    \textit{Anonymous Institution} \\
    City, Country \\
    anonymous@institution.example
  }
}

\else

% authors: me, Ralf, Ney

% Content for the camera-ready version
\author{\IEEEauthorblockN{Albert Zeyer \orcidlink{0000-0002-6655-671X}, Ralf Schlüter \orcidlink{0000-0003-2839-9247}, Hermann Ney}
\IEEEauthorblockA{\textit{Machine Learning and Human Language Technology Group, RWTH Aachen University, Aachen, Germany} \\
\textit{AppTek GmbH, Aachen, Germany}\\
% Aachen, Germany \\
\{zeyer,schlueter,ney\}@ml.rwth-aachen.de}
}

\fi

\maketitle

% *CRITICAL: Do Not Use Symbols, Special Characters, Footnotes,  or Math in Paper Title or Abstract.

\begin{abstract}
Speech-to-text alignment means finding the temporal boundaries of each word in the audio.
Some models provide such an alignment directly and others do not.
Connectionist temporal classification (CTC) and transducer models have an alignment by construction,
whereas attention-based encoder-decoders (AED) and speech large language models (LLMs) do not,
and their word timings are usually read off the attention weights instead.
All of these signals live on the encoder frame grid, which bounds their temporal precision.
We study a generic \emph{gradient-based alignment} that applies to any differentiable ASR model.
We take the gradient of each teacher-forced token log probability with respect to the input,
reduce it to a per-frame saliency,
and decode the resulting matrix into word boundaries with a single dynamic-programming pass.
The method needs no training, no model modification and no alignment heads,
works across all model families including the speech LLMs,
and aligns on the input grid rather than on the coarser encoder grid.
We evaluate it on sixteen models from four families,
on read (TIMIT) and spontaneous (Buckeye) speech,
each against the model's own native or attention-based alignment.
We find that the gradient yields a usable alignment for every model,
that it is usually somewhat behind a strong native aligner
but better where the native alignment is weak, as for the streaming models,
and that its main disadvantage is the cost of one backward pass per token.
\end{abstract}


\section{Introduction \& Related Work}

Speech-to-text alignment assigns each transcribed word a start and end time in the audio.
It is classically produced by finding the best path through
a Gaussian-mixture hidden Markov model (GM-HMM)
which is still the most accurate aligner for read speech~\cite{gales2008application,rabiner1989tutorial,mcauliffe2017montreal,huang2024lesspeaky,rousso24_interspeech}.
Current models, however, differ in what alignment they provide on their own~\cite{prabhavalkar2023end}:
connectionist temporal classification (CTC)~\cite{graves2006ctc,huang2024lesspeaky}
and transducer models~\cite{graves2012transduction} have one by construction,
whereas attention-based encoder-decoders (AED)~\cite{chorowski2015attention,chan2016las,zeyer2018:asr-attention}
and speech large language models (LLMs)~\cite{zhang2023speechgpt,chu2023qwenaudio,schmitt2026:speech-llm} do not,
and their word timings are usually read off the attention weights instead.
All of these signals live on the 20 to 80\,ms encoder frame grid, which bounds their temporal precision.
For AED, reading the timing off the cross-attention is established practice:
Whisper emits coarse segment-level timestamp tokens~\cite{radford2023whisper},
word-level timestamps follow from a dynamic time warping (DTW) of the decoder cross-attention~\cite{bain2023whisperx}
that fine-tuning can sharpen~\cite{wagner2024crisperwhisper},
and the alignment-bearing heads can even be selected by an unsupervised heuristic
and refined with character-level teacher forcing without any training~\cite{yeh2025whisperaligner}.
Speech LLM self-attention carries a similar alignment,
used to constrain text-to-speech synthesis~\cite{wang2024aeam}
and as a policy for simultaneous translation~\cite{papi2026doa};
forced alignment has separately been cast as a trained LLM task~\cite{mu2026llmforcedaligner}.

\begin{figure}
\begin{center}
\includegraphics[width=\linewidth]{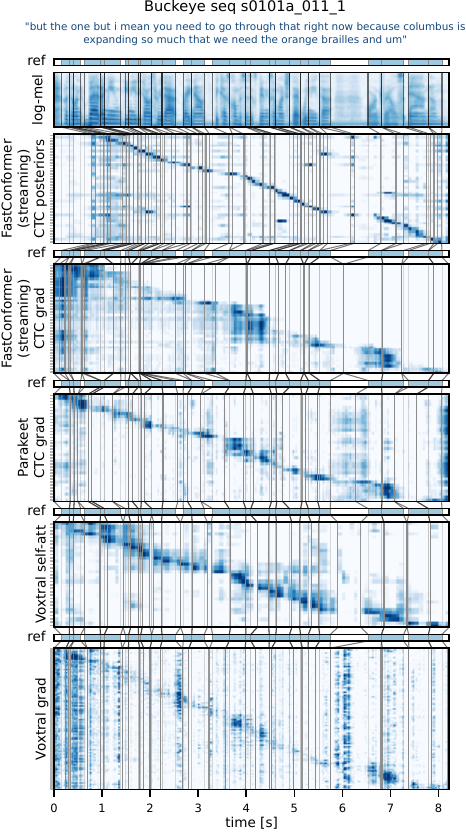}
\end{center}
\caption{Posteriors $\log \left( p_t(y{=}a_s \mid x_1^{T'}) \right)_{s,t} \in \R^{S \times T}$,
gradient scores $\log \operatorname{softmax}_{T} (G) \in \R^{S \times T}$,
and log self-attention weights (in $\R^{S \times T}$),
all without energy weighting here,
each with word boundaries,
in comparison to the reference segmentation (silence in white, words in blue, with word boundaries).
% the log softmax over time of the log gradient norm of $p(a_s \mid a_1^{s-1}, x_1^{T'})$ w.r.t.~the input frames $x_1^T$.
}
\label{fig:input-grad}
\end{figure}

We study a generic \emph{gradient-based alignment}
that applies to any differentiable speech recognition model.
For each transcript token we take the gradient of its teacher-forced log probability
with respect to the input,
reduce it to a per-frame saliency~\cite{simonyan2014deep},
and decode the resulting token-by-frame matrix into word boundaries
with a single dynamic-programming pass (\cref{fig:input-grad}).
As it uses the gradient w.r.t.~the input signal,
it aligns on the input grid rather than on the coarser encoder grid,
and it can correct temporal shifts of the encoder:
the encoder is often so powerful that it can displace the signal in time (e.g.~with streaming models)
or even reverse the time dimension~\cite{schmitt2025:flipped-conformer}%
\footnote{Arguably reversing the time dimension will not happen for CTC, though.},
which degrades the native forced alignment,
while the gradient should still give a meaningful alignment.
It needs no training and no model modification
and it applies to all model families.
Unlike the attention-based alignment,
which first has to find which attention head carries the alignment
(hand-picked for Whisper, and unpublished for the other models,
so that we have to select it on a labeled development set),
the gradient uses one canonical signal, the saliency of the token's own log probability.
In prior work, the same method has been applied to speech recognition AED models~\cite{schmitt2025:flipped-conformer,yeh2025whisperaligner}
% yeh2025whisperaligner: only on encoder output
and machine-translation models~\cite{ding2019saliency}.

This work is not a proposal of gradient alignment as the best aligner.
A strong native aligner is usually still somewhat better, and the gradient is considerably more expensive.
We rather provide a broad and fair analysis of what the gradient alignment is,
how well it works across all the model families, and where it wins and loses.
Our contributions are:
\begin{itemize}
\item We show that the gradient-based alignment applies to all ASR model families:
CTC (prefix scores), transducers (RNN-T and TDT, also via prefix scores),
AED, and speech LLMs.
\item Improved alignment path scoring, where the best alignment path is found via dynamic programming.
The cross-attention dynamic time warping (DTW) used by Whisper and CrisperWhisper is a special case of it.
Our version improves on that DTW both on the attention and on the gradient signal.
\item For the speech LLMs, we read the alignment off their self-attention,
  the analog of the encoder-decoder cross-attention.
\item A comprehensive and fair comparison of sixteen models across the four families,
  on read (TIMIT) and spontaneous (Buckeye) speech,
  each against its own native or attention-based alignment,
  including the tokenization granularity, the input-grid resolution,
  the recognition-mode alignment, and the compute cost.
\item Public source code to reproduce all results including the whole pipeline.
\end{itemize}

\section{Alignments via Gradients}

We can calculate the gradient of the log probability $p(a_s \mid a_1^{s-1}, x_1^{T'})$
of some target label $a_s \in \mathcal{A}$ (the label vocabulary) in label position $s$,
given the input $x_1^{T'}$ and the history $a_1^{s-1} \in \mathcal{A}^{s-1}$ of previous labels,
w.r.t.~an input frame $x_t$.
The transcript $a_1^S$ has $S$ labels and the input $T'$ frames;
the saliency matrix and the alignment below have $T$ time frames:
$T = T'$ when the gradient is taken at the input,
or the number of encoder frames when it is taken at an encoder layer.
Comparing the norm of these gradients over the time frames $t$
will give us an indication of the importance of each frame for this specific output label $a_s$.
Specifically, we calculate the log norm%
\footnote{We found that the log norm was better conditioned than the norm, and yielded better results.
We also tested different $p$-norms, and found $p=2$ in most cases to perform best.}
\begin{equation}
G_{s,t} := \log \left\Vert \nabla_{x_t} \log p(a_s \mid a_1^{s-1}, x_1^{T'}) \right\Vert_p \in \mathbb{R}.
\label{eq:grad}
\end{equation}
The matrix $\log \operatorname{softmax}_{T} G$ in \Cref{fig:input-grad} shows the alignment clearly.

% blankStopGrad-inclBlankState-p0.1
% -> include_next_blank=True
Note that $\log p(a_s \mid a_1^{s-1}, x_1^{T'})$
is straightforward to compute for an AED model or speech LLM (we exclude the EOS label here),
and was done in a similar way in \cite{schmitt2025:flipped-conformer,yeh2025whisperaligner}.
It is possible for CTC and transducer as well,  % probably also HMM, but don't mention now...
using the prefix scores \cite{hori-etal-2017-joint}
%$\log p(\overline{a}_s \mid \overline{a}_1^{s-1}, x_1^{T'})$
%\begin{align}
%\log p_{\textrm{CTC}}(\overline{a}_s \mid \overline{a}_1^{s-1}, x_1^{T'})
%= \log \sum_{t \le T, y_1^t \colon a_1^s} p(y_1^t \mid h_1^t)
%- \log \sum_{t \le T, y_1^t \colon a_1^{s-1}} p(y_1^t \mid h_1^t) .
%\end{align}
which can be calculated efficiently using dynamic programming.%
%\footnote{In fact, calculating the prefix scores is already part of the usual CTC loss calculation itself.}.

% (blankstop / stop_grad_blank, seems that it now does not help anymore in our new setup, so let's ignore it.)
%As a further tweak, we slightly modify the gradients of the logits by masking out the gradients of the blank logit. I.e.~in the automatic differentiation, we hook after the gradient computation of $\nabla_z L$, and then
%\begin{align}
%$(\nabla_z L)_{\blank} \leftarrow 0$.
%\end{align}
%This slightly improves our results. %(see \Cref{sec:appendix:grad-align:blank-stop-grad}).

To use this to get some alignment,
we need to define which alignment label topology we allow (mapping $a_1^S$ to $y_1^T$)
and how to score one particular alignment $y_1^T$
such that we can search for the one with the highest score.

\begin{figure}[t]
\centering
\begin{tikzpicture}[
  x=0.52cm, y=0.42cm, font=\footnotesize,
  st/.style={circle, fill=black, inner sep=1.4pt},
  mv/.style={-{Stealth[length=4pt]}, thick},
]
\def\T{11}\def\S{7}
% forbidden intra-word blank state (row 3): shaded / disallowed
\fill[black!8] (0.5,2.55) rectangle (\T+0.5,3.45);
\foreach \t in {1,...,\T}{\draw[black!12] (\t,0.6)--(\t,\S+0.4);}
\foreach \s in {1,...,\S}{\draw[black!12] (0.6,\s)--(\T+0.4,\s);}
\foreach \s/\lab in {1/{$\blank$},2/{$a_1$},3/{$\blank$},4/{$a_2$},5/{$\blank$},6/{$a_3$},7/{$\blank$}}{
  \node[left, inner sep=1pt] at (0.5,\s) {\lab};}
\node[black!55, font=\scriptsize] at (8.5,3) {no intra-word $\blank$};
\node[font=\scriptsize] at (\T/2+0.5,-0.2) {time $t$};
% Viterbi path
\node[st](a) at (1,1){}; \node[st](b) at (2,2){}; \node[st](c) at (3,2){};
\node[st](d) at (4,2){}; \node[st](e) at (5,4){}; \node[st](f) at (6,4){};
\node[st](g) at (7,4){}; \node[st](h) at (8,5){}; \node[st](i) at (9,6){};
\node[st](j) at (10,6){}; \node[st](k) at (11,7){};
\draw[mv](a)--(b); \draw[mv](b)--(c); \draw[mv](c)--(d); \draw[mv](d)--(e); \draw[mv](e)--(f); \draw[mv](f)--(g); \draw[mv](g)--(h); \draw[mv](h)--(i); \draw[mv](i)--(j); \draw[mv](j)--(k);
\node[font=\scriptsize, below, inner sep=1.5pt] at (3,2){stay};
\node[font=\scriptsize] at (4.55,3.55){skip $\blank$};
\end{tikzpicture}
\caption{Decoding the saliency into an alignment, for two words $(a_1 a_2)(a_3)$.
The DP runs a time-synchronous Viterbi over the FSA states $Y_1^{2S+1}$ (rows): per frame it stays,
advances one state ($+1$), or skips a blank.
The word-level topology shades out the intra-word blank
(between $a_1$ and $a_2$), forcing a skip there, while keeping the word-boundary blank (state $5$).
DTW instead allows a vertical move (advancing the token within a frame),
so adjacent tokens may overlap by up to one frame.}
\label{fig:dp-grid}
\end{figure}
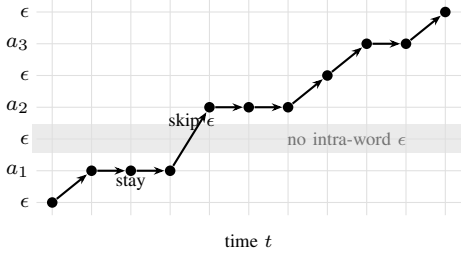

For the label topology, we map $a_1^S$ to $y_1^T$ over $\mathcal{Y} = \mathcal{A} \cup \{\blank\}$,
letting each real label repeat over consecutive frames with $\blank$ (blank/silence) labels in between.
We write this as a finite state automaton
with enumerated states $Y_1^{2S+1} = (\blank, 1, \blank, 2, \dots, S, \blank)$,
whose $S+1$ blank states sit before the first label, between adjacent labels, and after the last.
A topology is fixed by which of these blank states it permits, the rest being disallowed.
Our default is a \emph{word-level} topology: a blank only at word boundaries
(before the first word, after the last, and between adjacent words), forbidding blanks inside a word,
as we evaluate word boundaries.
The \emph{full}, CTC-like topology instead permits every blank
(unlike CTC, we do not force a $\blank$ between two equal labels $a_s = a_{s+1}$),
and a variant with \emph{no interior silence} keeps only the leading and trailing blank,
so adjacent tokens follow each other directly.
We decode all topologies with the standard time-synchronous Viterbi search (as in CTC),
advancing one frame per step, so every token spans at least one frame.
Whisper and CrisperWhisper instead align the cross-attention by DTW,
which additionally allows vertical transitions (advancing the token within a frame),
so adjacent tokens may overlap by up to one frame;
with no interior silence and without the log of \cref{eq:grad-token-score},
our decoder reduces to that DTW up to those vertical transitions (\cref{tab:alignopts-dtw}).
We search for an allowed state sequence $r_1^T$,
$r_t \in \{1,\dots,2S+1\}$, compatible with $a_1^S$, that maximizes
\begin{equation}
\operatorname{GradScore}(r_1^T) = \sum_{t=1}^T
\begin{cases}
G'_{Y_{r_t},t}, & Y_{r_t} \ne \blank, \\
\beta_t, & Y_{r_t} = \blank ,
\end{cases}
\label{eq:grad-score}
\end{equation}
with a token score $G'$ and a silence score $\beta_t$ defined below.
The best $r_1^T$ is found via dynamic programming (\cref{fig:dp-grid}),
and the alignment $y_1^T$ read off with
$y_t = a_{Y_{r_t}}$ for non-blank states and $y_t = \blank$ otherwise.

The token score is the over-time log-softmax of the (optionally energy-weighted) saliency,
\begin{equation}
G'_{s,t} = \log\operatorname{softmax}_{T} (G + \rho \log E)_{s,t} ,
\label{eq:grad-token-score}
\end{equation}
where $\rho \ge 0$ is a weighting ($\rho = 0.5$ by default; $\rho = 0$ disables it)
and $E_t \in [0,1]$ a smoothed audio-energy envelope on the frame grid:
from the waveform $x$ we form the windowed root-mean-square energy $\tilde{E} = \sqrt{w * x^2}$,
with $w$ a normalized $25$\,ms Hann window ($\sum_i w_i = 1$) and $*$ convolution,
sample it at the frame centers $c_t$, and normalize by its maximum,
$E_t = \tilde{E}_{c_t} / \max_{t'} \tilde{E}_{c_{t'}}$.
The energy weight suppresses the gradient's spurious response in silent frames.

The silence score $\beta_t$ admits the constant blank of CTC,
but also a self-calibrating blank derived from the per-frame statistics of the token scores.
Let $\mu_t$ and $\sigma_t$ be the mean and standard deviation of $\{G'_{s,t}\}_s$ over the labels $s$
(taken at $\rho = 0$, so the blank calibrates against the raw saliency).
We consider
\begin{equation}
\beta_t =
\begin{cases}
\gamma, & \text{(constant)} \\
\mu_t + \kappa\,\sigma_t, & \text{(z-score)} \\
\mu_t - \lambda\,z(E_t)\,\sigma_t, & \text{(energy)}
\end{cases}
\label{eq:silence-score}
\end{equation}
with $z(E_t) = (E_t - \bar{E})/\operatorname{std}(E)$ the z-score of the energy over the $T$ frames
($\bar{E}$ and $\operatorname{std}(E)$ its mean and standard deviation),
and hyperparameters $\gamma$ (constant level), $\kappa$ and $\lambda$.
The energy blank is a VAD-style silence emission:
in a low-energy frame $z(E_t) < 0$ raises $\beta_t$ above the token mean, so the frame goes to the blank;
in speech frames $\beta_t$ drops below the token scores.

\begin{table}[tb]
  \centering
  \caption{%
\textbf{Word-boundary error and accuracy per model},
gradient alignment vs.~each model's native or attention aligner,
on TIMIT-test and Buckeye, grouped by family.
The attention and posterior aligners use the model's native subword units;
the gradient uses characters for AED and speech-LLM
and native subwords otherwise (\cref{tab:char-vs-subword-family}).
Starred Whisper-large-v3 gradient row (*) is taken at optimal encoder depth (encoder 3/4, \cref{tab:encoder-depth}).%
}

  \label{tab:per-model-methods}
  \begin{adjustbox}{max width=\linewidth}
  \begin{tabular}{|l|l|l|r|r|r|r|}
    \hline
    \multicolumn{2}{|c|}{\textbf{Model}} & \multirow{2}{*}{\textbf{\makecell[l]{Align\\method}}} & \multicolumn{2}{c|}{\textbf{TIMIT}} & \multicolumn{2}{c|}{\textbf{Buckeye}} \\
    \cline{1-2} \cline{4-5} \cline{6-7}
    \textbf{Type} & \textbf{Name} &  & \textbf{\makecell{WBE \\ {\normalfont {[ms]}\,$\downarrow$}}} & \textbf{\makecell{$\le$50ms \\ {\normalfont {[\%]}\,$\uparrow$}}} & \textbf{\makecell{WBE \\ {\normalfont {[ms]}\,$\downarrow$}}} & \textbf{\makecell{$\le$50ms \\ {\normalfont {[\%]}\,$\uparrow$}}} \\ \hline\hline
    % (cont.): t_wbe,t_a50 <- baseline-mfa-timit-test-wbe.txt  |  s_wbe,s_a50 <- baseline-mfa-buckeye-segA-5h-wbe.txt
    \makecell[l]{GM-\\HMM} & MFA & Likelihood & $19$ & $92.3$ & $32$ & $90.0$ \\
    \hline\hline
    % (cont.): t_wbe,t_a50 <- align/wav2vec2ctc-fproj_out-prefixfwd-timit-test-L2_grad-pertoken-asotTrue-bs-5-en0.5-sil2.0-wordtopo-wbe.txt  |  s_wbe,s_a50 <- align/wav2vec2ctc-fproj_out-prefixfwd-buckeye-segA-5h-L2_grad-pertoken-asotTrue-bs-5-en0.5-sil2.0-wordtopo-wbe.txt
    \multirow{10}{*}{CTC} & \multirow{2}{*}{MMS-FA} & Gradients & $49$ & $64.3$ & $121$ & $58.2$ \\
    \cline{3-7}
    % (cont.): t_wbe,t_a50 <- baseline-mms_fa-timit-test-wbe.txt  |  s_wbe,s_a50 <- baseline-mms_fa-buckeye-segA-5h-wbe.txt
     &  & Posteriors & $37$ & $71.5$ & $46$ & $69.9$ \\
    \cline{2-7}
    % (cont.): t_wbe,t_a50 <- align/phoneme-vitouphy-prefixfwd-timit-test-L2_grad-pertoken-g2pword-asotTrue-bs-5-en0.5-sil2.0-wordtopo-wbe.txt  |  s_wbe,s_a50 <- align/phoneme-vitouphy-prefixfwd-buckeye-segA-5h-L2_grad-pertoken-g2pword-asotTrue-bs-5-en0.5-sil2.0-wordtopo-wbe.txt
     & \multirow{2}{*}{\makecell[l]{XLS-R \\ (Phoneme)}} & Gradients & $49$ & $65.7$ & $81$ & $66.4$ \\
    \cline{3-7}
    % (cont.): t_wbe,t_a50 <- align/w2v-phoneme-timit-test-ctc-forced-align-word-wbe.txt  |  s_wbe,s_a50 <- baseline-phoneme-fa-buckeye-segA-5h-word-wbe.txt
     &  & Posteriors & $30$ & $80.7$ & $44$ & $70.4$ \\
    \cline{2-7}
    % (cont.): t_wbe,t_a50 <- align/parakeet-ctc-1.1b-prefixfwd-timit-test-L2_grad-pertoken-asotTrue-bs-5-en0.5-sil2.0-wordtopo-wbe.txt  |  s_wbe,s_a50 <- align/parakeet-ctc-1.1b-prefixfwd-buckeye-segA-5h-L2_grad-pertoken-asotTrue-bs-5-en0.5-sil2.0-wordtopo-wbe.txt
     & \multirow{2}{*}{\makecell[l]{Parakeet\\CTC}} & Gradients & $94$ & $40.8$ & $117$ & $37.1$ \\
    \cline{3-7}
    % (cont.): t_wbe,t_a50 <- baseline-parakeet-ctc-1.1b-timit-test-wbe.txt  |  s_wbe,s_a50 <- baseline-parakeet-ctc-1.1b-buckeye-segA-5h-wbe.txt
     &  & Posteriors & $77$ & $39.4$ & $99$ & $32.2$ \\
    \cline{2-7}
    % (cont.): t_wbe,t_a50 <- align/owsm-ctc-v4-1b-lyr6-timit-test-L2_grad-pertoken-asotTrue-bs-5-en0.5-sil2.0-wordtopo-wbe.txt  |  s_wbe,s_a50 <- align/owsm-ctc-v4-1b-lyr6-buckeye-segA-5h-L2_grad-pertoken-asotTrue-bs-5-en0.5-sil2.0-wordtopo-wbe.txt
     & \multirow{2}{*}{\makecell[l]{OWSM-\\CTC}} & Gradients & $142$ & $28.4$ & $145$ & $30.4$ \\
    \cline{3-7}
    % (cont.): t_wbe,t_a50 <- baseline-owsm-ctc-v4-1b-lyr6-timit-test-wbe.txt  |  s_wbe,s_a50 <- baseline-owsm-ctc-v4-1b-lyr6-buckeye-segA-5h-wbe.txt
     &  & Posteriors & $120$ & $20.6$ & $110$ & $25.2$ \\
    \cline{2-7}
    % (cont.): t_wbe,t_a50 <- align/fastconformer-stream-ctc-timit-test-L2_grad-pertoken-asotTrue-bs-5-en0.5-sil2.0-wordtopo-wbe.txt  |  s_wbe,s_a50 <- align/fastconformer-stream-ctc-buckeye-segA-5h-L2_grad-pertoken-asotTrue-bs-5-en0.5-sil2.0-wordtopo-wbe.txt
     & \multirow{2}{*}{\makecell[l]{FastConformer\\(streaming)}} & Gradients & $127$ & $32.0$ & $158$ & $30.1$ \\
    \cline{3-7}
    % (cont.): t_wbe,t_a50 <- baseline-fastconformer-stream-ctc-timit-test-wbe.txt  |  s_wbe,s_a50 <- baseline-fastconformer-stream-ctc-buckeye-segA-5h-wbe.txt
     &  & Posteriors & $357$ & $1.1$ & $366$ & $2.2$ \\
    \hline\hline
    % (cont.): t_wbe,t_a50 <- align/parakeet-rnnt-1.1b-logmel-timit-test-L2_grad-pertoken-asotTrue-bs-5-en0.5-sil2.0-wordtopo-wbe.txt  |  s_wbe,s_a50 <- align/parakeet-rnnt-1.1b-logmel-buckeye-segA-5h-L2_grad-pertoken-asotTrue-bs-5-en0.5-sil2.0-wordtopo-wbe.txt
    \multirow{8}{*}{Transd.} & \multirow{2}{*}{\makecell[l]{Parakeet\\RNN-T}} & Gradients & $125$ & $31.5$ & $147$ & $29.1$ \\
    \cline{3-7}
    % (cont.): t_wbe,t_a50 <- baseline-parakeet-rnnt-1.1b-native-viterbi-timit-test-wbe.txt  |  s_wbe,s_a50 <- baseline-parakeet-rnnt-1.1b-native-viterbi-buckeye-segA-5h-wbe.txt
     &  & Posteriors & $79$ & $39.3$ & $93$ & $35.4$ \\
    \cline{2-7}
    % (cont.): t_wbe,t_a50 <- align/parakeet-tdt-0.6b-v2-logmel-timit-test-L2_grad-pertoken-asotTrue-bs-5-en0.5-sil2.0-wordtopo-wbe.txt  |  s_wbe,s_a50 <- align/parakeet-tdt-0.6b-v2-logmel-buckeye-segA-5h-L2_grad-pertoken-asotTrue-bs-5-en0.5-sil2.0-wordtopo-wbe.txt
     & \multirow{2}{*}{\makecell[l]{Parakeet\\TDT}} & Gradients & $128$ & $32.9$ & $157$ & $27.5$ \\
    \cline{3-7}
    % (cont.): t_wbe,t_a50 <- baseline-parakeet-tdt-0.6b-v2-native-viterbi-timit-test-wbe.txt  |  s_wbe,s_a50 <- baseline-parakeet-tdt-0.6b-v2-native-viterbi-buckeye-segA-5h-wbe.txt
     &  & Posteriors & $80$ & $41.2$ & $90$ & $38.6$ \\
    \cline{2-7}
    % (cont.): t_wbe,t_a50 <- align/emformer-rnnt-prefix-logmel-timit-test-L2_grad-pertoken-asotTrue-bs-5-en0.5-sil2.0-wordtopo-wbe.txt  |  s_wbe,s_a50 <- align/emformer-rnnt-prefix-logmel-buckeye-segA-5h-L2_grad-pertoken-asotTrue-bs-5-en0.5-sil2.0-wordtopo-wbe.txt
     & \multirow{2}{*}{\makecell[l]{Emformer\\(streaming)}} & Gradients & $159$ & $24.8$ & $139$ & $30.4$ \\
    \cline{3-7}
    % (cont.): t_wbe,t_a50 <- baseline-emformer-rnnt-native-viterbi-timit-test-wbe.txt  |  s_wbe,s_a50 <- baseline-emformer-rnnt-native-viterbi-buckeye-segA-5h-wbe.txt
     &  & Posteriors & $394$ & $0.3$ & $399$ & $0.5$ \\
    \cline{2-7}
    % (cont.): t_wbe,t_a50 <- align/fastconformer-stream-rnnt-timit-test-L2_grad-pertoken-asotTrue-bs-5-en0.5-sil2.0-wordtopo-wbe.txt  |  s_wbe,s_a50 <- align/fastconformer-stream-rnnt-buckeye-segA-5h-L2_grad-pertoken-asotTrue-bs-5-en0.5-sil2.0-wordtopo-wbe.txt
     & \multirow{2}{*}{\makecell[l]{FastConformer\\(streaming)}} & Gradients & $153$ & $27.3$ & $189$ & $24.0$ \\
    \cline{3-7}
    % (cont.): t_wbe,t_a50 <- baseline-fastconformer-stream-rnnt-native-viterbi-timit-test-wbe.txt  |  s_wbe,s_a50 <- baseline-fastconformer-stream-rnnt-native-viterbi-buckeye-segA-5h-wbe.txt
     &  & Posteriors & $307$ & $3.6$ & $331$ & $3.9$ \\
    \hline\hline
    % (cont.): t_wbe,t_a50 <- align/whisper-base-logmel-timit-test-L2_grad-pertoken-charlev-spc-asotTrue-bs-5-en0.5-sil2.0-wordtopo-wbe.txt  |  s_wbe,s_a50 <- align/whisper-base-logmel-buckeye-segA-5h-L2_grad-pertoken-charlev-spc-asotTrue-bs-5-en0.5-sil2.0-wordtopo-wbe.txt
    \multirow{9}{*}{AED} & \multirow{2}{*}{\makecell[l]{Whisper-\\base}} & Gradients & $51$ & $59.3$ & $75$ & $52.7$ \\
    \cline{3-7}
    % (cont.): t_wbe,t_a50 <- align/baseline-whisper-base-crossattn-auto-timit-test-asotTrue-bs-5-en0.5-sil2.0-wordtopo-wbe.txt  |  s_wbe,s_a50 <- align/baseline-whisper-base-crossattn-auto-buckeye-segA-5h-asotTrue-bs-5-en0.5-sil2.0-wordtopo-wbe.txt
     &  & Cross-att. & $51$ & $62.3$ & $62$ & $66.3$ \\
    \cline{2-7}
    % (cont.): t_wbe,t_a50 <- align/whisper-large-v3-logmel-timit-test-L2_grad-pertoken-charlev-spc-asotTrue-bs-5-en0.5-sil2.0-wordtopo-wbe.txt  |  s_wbe,s_a50 <- align/whisper-large-v3-logmel-buckeye-segA-5h-L2_grad-pertoken-charlev-spc-asotTrue-bs-5-en0.5-sil2.0-wordtopo-wbe.txt
     & \multirow{3}{*}{\makecell[l]{Whisper-\\large-v3}} & Gradients & $55$ & $57.8$ & $53$ & $66.2$ \\
    \cline{3-7}
    % (cont.): t_wbe,t_a50 <- align/whisper-large-v3-charlev-spc-encL24-encdepth-timit-test-L2_grad-pertoken-asotTrue-bs-5-en0.5-sil2.0-wordtopo-wbe.txt  |  s_wbe,s_a50 <- align/whisper-large-v3-charlev-spc-encL24-encdepth-buckeye-segA-5h-L2_grad-pertoken-asotTrue-bs-5-en0.5-sil2.0-wordtopo-wbe.txt
     &  & Gradients* & $33$ & $76.9$ & $39$ & $80.0$ \\
    \cline{3-7}
    % (cont.): t_wbe,t_a50 <- align/baseline-whisper-large-v3-crossattn-auto-timit-test-asotTrue-bs-5-en0.5-sil2.0-wordtopo-wbe.txt  |  s_wbe,s_a50 <- align/baseline-whisper-large-v3-crossattn-auto-buckeye-segA-5h-asotTrue-bs-5-en0.5-sil2.0-wordtopo-wbe.txt
     &  & Cross-att. & $42$ & $66.7$ & $49$ & $69.1$ \\
    \cline{2-7}
    % (cont.): t_wbe,t_a50 <- align/crisperwhisper-logmel-timit-test-L2_grad-pertoken-charlev-spc-asotTrue-bs-5-en0.5-sil2.0-wordtopo-wbe.txt  |  s_wbe,s_a50 <- align/crisperwhisper-logmel-buckeye-segA-5h-L2_grad-pertoken-charlev-spc-asotTrue-bs-5-en0.5-sil2.0-wordtopo-wbe.txt
     & \multirow{2}{*}{\makecell[l]{Crisper-\\Whisper}} & Gradients & $60$ & $54.3$ & $59$ & $59.2$ \\
    \cline{3-7}
    % (cont.): t_wbe,t_a50 <- align/baseline-crisperwhisper-crossattn-auto-timit-test-asotTrue-bs-5-en0.5-sil2.0-wordtopo-wbe.txt  |  s_wbe,s_a50 <- align/baseline-crisperwhisper-crossattn-auto-buckeye-segA-5h-asotTrue-bs-5-en0.5-sil2.0-wordtopo-wbe.txt
     &  & Cross-att. & $33$ & $80.7$ & $47$ & $80.1$ \\
    \cline{2-7}
    % (cont.): t_wbe,t_a50 <- align/owls-1B-180K-charlev-logmel-timit-test-L2_grad-pertoken-asotTrue-bs-5-en0.5-sil2.0-wordtopo-wbe.txt  |  s_wbe,s_a50 <- align/owls-1B-180K-charlev-logmel-buckeye-segA-5h-L2_grad-pertoken-asotTrue-bs-5-en0.5-sil2.0-wordtopo-wbe.txt
     & \multirow{2}{*}{OWLS-1B} & Gradients & $153$ & $31.7$ & $180$ & $32.9$ \\
    \cline{3-7}
    % (cont.): t_wbe,t_a50 <- align/baseline-owls-1B-180K-crossattn-auto-timit-test-asotTrue-bs-5-en0.5-sil2.0-wordtopo-wbe.txt  |  s_wbe,s_a50 <- align/baseline-owls-1B-180K-crossattn-auto-buckeye-segA-5h-asotTrue-bs-5-en0.5-sil2.0-wordtopo-wbe.txt
     &  & Cross-att. & $35$ & $76.7$ & $51$ & $69.1$ \\
    \hline\hline
    % (cont.): t_wbe,t_a50 <- align/voxtral-charlevlogmel-timit-test-L2_grad-pertoken-asotTrue-bs-5-en0.5-sil2.0-wordtopo-wbe.txt  |  s_wbe,s_a50 <- align/voxtral-charlevlogmel-buckeye-segA-5h-L2_grad-pertoken-asotTrue-bs-5-en0.5-sil2.0-wordtopo-wbe.txt
    \multirow{6}{*}{\makecell[l]{Speech \\ LLM}} & \multirow{2}{*}{Voxtral} & Gradients & $77$ & $45.6$ & $74$ & $51.0$ \\
    \cline{3-7}
    % (cont.): t_wbe,t_a50 <- align/baseline-voxtral-selfattn-timit-test-asotTrue-bs-5-en0.5-sil2.0-wordtopo-wbe.txt  |  s_wbe,s_a50 <- align/baseline-voxtral-selfattn-buckeye-segA-5h-asotTrue-bs-5-en0.5-sil2.0-wordtopo-wbe.txt
     &  & Self-att. & $53$ & $61.1$ & $52$ & $65.5$ \\
    \cline{2-7}
    % (cont.): t_wbe,t_a50 <- align/phi4mm-timit-test-L2_grad-pertoken-charlev-spc-asotTrue-bs-5-en0.5-sil2.0-wordtopo-wbe.txt  |  s_wbe,s_a50 <- align/phi4mm-buckeye-segA-5h-L2_grad-pertoken-charlev-spc-asotTrue-bs-5-en0.5-sil2.0-wordtopo-wbe.txt
     & \multirow{2}{*}{Phi-4-MM} & Gradients & $84$ & $43.0$ & $104$ & $43.1$ \\
    \cline{3-7}
    % (cont.): t_wbe,t_a50 <- align/baseline-phi4mm-selfattn-timit-test-asotTrue-bs-5-en0.5-sil2.0-wordtopo-wbe.txt  |  s_wbe,s_a50 <- align/baseline-phi4mm-selfattn-buckeye-segA-5h-asotTrue-bs-5-en0.5-sil2.0-wordtopo-wbe.txt
     &  & Self-att. & $68$ & $46.6$ & $79$ & $39.8$ \\
    \cline{2-7}
    % (cont.): t_wbe,t_a50 <- align/canary-qwen-charlev-spc-logmel-st15-timit-test-L2_grad-pertoken-asotTrue-bs-5-en0.5-sil2.0-wordtopo-wbe.txt  |  s_wbe,s_a50 <- align/canary-qwen-charlev-spc-logmel-st15-buckeye-segA-5h-L2_grad-pertoken-asotTrue-bs-5-en0.5-sil2.0-wordtopo-wbe.txt
     & \multirow{2}{*}{Canary-Qwen} & Gradients & $90$ & $40.6$ & $99$ & $40.6$ \\
    \cline{3-7}
    % (cont.): t_wbe,t_a50 <- align/baseline-canary-qwen-selfattn-timit-test-asotTrue-bs-5-en0.5-sil2.0-wordtopo-wbe.txt  |  s_wbe,s_a50 <- align/baseline-canary-qwen-selfattn-buckeye-segA-5h-asotTrue-bs-5-en0.5-sil2.0-wordtopo-wbe.txt
     &  & Self-att. & $131$ & $28.0$ & $211$ & $16.3$ \\
    \hline
  \end{tabular}
  \end{adjustbox}
\end{table}

% =====================================================================
% AUTO-GENERATED by scripts/render_tables.py -- DO NOT EDIT THIS FILE.
% Regenerate from (edit these, then re-run render_tables.py / sync_tables_preview.sh):
%   tables-data/hyp.data.json    -- the numbers (from the Sisyphus graph)
%   tables-spec/hyp.spec.json     -- columns, units, layout
%   tables-spec/hyp.caption.tex   -- caption text
% =====================================================================
% data source (Sisyphus recipe): i6_experiments/users/zeyer/experiments/exp2025_07_07_in_grads/jobs/grad_align_tables.py :: _hyp_table
% per-row comments below map columns -> the registered output name the number comes from
\begin{table}[tb]
  \centering
  \caption{%
\textbf{Hypothesis-mode alignment} on Buckeye, where each model aligns its own recognition.
Identity-gated F1 at a $50$\,ms collar and Levenshtein-matched WBE against the reference.%
}

  \label{tab:hyp}
  \begin{adjustbox}{max width=\linewidth}
  \begin{tabular}{|l|l|r|r|l|r|r|}
    \hline
    \multicolumn{2}{|c|}{\textbf{Model}} & \textbf{\makecell{Ref-}} & \multirow{2}{*}{\textbf{\makecell{WER \\ {\normalfont {[\%]}\,$\downarrow$}}}} & \multirow{2}{*}{\textbf{\makecell[l]{Align\\method}}} & \textbf{\makecell{Matched-}} & \textbf{\makecell{F1}} \\
    \cline{1-2}
    \textbf{Type} & \textbf{Name} & \textbf{\makecell{match \\ {\normalfont {[\%]}\,$\uparrow$}}} &  &  & \textbf{\makecell{WBE \\ {\normalfont {[ms]}\,$\downarrow$}}} & \textbf{\makecell{$\le$50ms \\ {\normalfont {[\%]}\,$\uparrow$}}} \\ \hline\hline
    % (cont.): rm,wer,mwbe,f50 <- hyp-align/parakeet-ctc-1.1b-buckeye-segA-5h-grad-metrics.txt
    \multirow{6}{*}{CTC} & \multirow{2}{*}{\makecell[l]{Parakeet\\CTC}} & \multirow{2}{*}{$92.3$} & \multirow{2}{*}{$11.9$} & Gradients & $116$ & $14.0$ \\
    \cline{5-7}
    % (cont.): rm,wer,mwbe,f50 <- hyp-align/parakeet-ctc-1.1b-buckeye-segA-5h-native-metrics.txt
     &  &  &  & Posteriors & $99$ & $11.1$ \\
    \cline{2-7}
    % (cont.): rm,wer,mwbe,f50 <- hyp-align/owsm-ctc-v4-1b-buckeye-segA-5h-grad-metrics.txt
     & \multirow{2}{*}{\makecell[l]{OWSM-\\CTC}} & \multirow{2}{*}{$89.2$} & \multirow{2}{*}{$14.3$} & Gradients & $142$ & $9.9$ \\
    \cline{5-7}
    % (cont.): rm,wer,mwbe,f50 <- hyp-align/owsm-ctc-v4-1b-buckeye-segA-5h-native-metrics.txt
     &  &  &  & Posteriors & $114$ & $5.4$ \\
    \cline{2-7}
    % (cont.): rm,wer,mwbe,f50 <- hyp-align/fastconformer-stream-ctc-buckeye-segA-5h-grad-metrics.txt
     & \multirow{2}{*}{\makecell[l]{FastConformer\\(streaming)}} & \multirow{2}{*}{$86.5$} & \multirow{2}{*}{$16.8$} & Gradients & $164$ & $9.7$ \\
    \cline{5-7}
    % (cont.): rm,wer,mwbe,f50 <- hyp-align/fastconformer-stream-ctc-buckeye-segA-5h-native-metrics.txt
     &  &  &  & Posteriors & $363$ & $0.2$ \\
    \hline\hline
    % (cont.): rm,wer,mwbe,f50 <- hyp-align/parakeet-rnnt-1.1b-logmel-buckeye-segA-5h-grad-metrics.txt
    \multirow{8}{*}{Transd.} & \multirow{2}{*}{\makecell[l]{Parakeet\\RNN-T}} & \multirow{2}{*}{$89.8$} & \multirow{2}{*}{$13.6$} & Gradients & $156$ & $7.6$ \\
    \cline{5-7}
    % (cont.): rm,wer,mwbe,f50 <- hyp-align/parakeet-rnnt-1.1b-logmel-buckeye-segA-5h-native-metrics.txt
     &  &  &  & Posteriors & $92$ & $13.2$ \\
    \cline{2-7}
    % (cont.): rm,wer,mwbe,f50 <- hyp-align/parakeet-tdt-0.6b-v2-logmel-buckeye-segA-5h-grad-metrics.txt
     & \multirow{2}{*}{\makecell[l]{Parakeet\\TDT}} & \multirow{2}{*}{$91.8$} & \multirow{2}{*}{$12.8$} & Gradients & $155$ & $7.8$ \\
    \cline{5-7}
    % (cont.): rm,wer,mwbe,f50 <- hyp-align/parakeet-tdt-0.6b-v2-logmel-buckeye-segA-5h-native-metrics.txt
     &  &  &  & Posteriors & $86$ & $17.1$ \\
    \cline{2-7}
    % (cont.): rm,wer,mwbe,f50 <- hyp-align/emformer-rnnt-prefix-logmel-buckeye-segA-5h-grad-metrics.txt
     & \multirow{2}{*}{\makecell[l]{Emformer\\(streaming)}} & \multirow{2}{*}{$72.1$} & \multirow{2}{*}{$30.5$} & Gradients & $155$ & $7.1$ \\
    \cline{5-7}
    % (cont.): rm,wer,mwbe,f50 <- hyp-align/emformer-rnnt-prefix-logmel-buckeye-segA-5h-native-metrics.txt
     &  &  &  & Posteriors & $391$ & $0.0$ \\
    \cline{2-7}
    % (cont.): rm,wer,mwbe,f50 <- hyp-align/fastconformer-stream-rnnt-buckeye-segA-5h-grad-metrics.txt
     & \multirow{2}{*}{\makecell[l]{FastConformer\\(streaming)}} & \multirow{2}{*}{$87.1$} & \multirow{2}{*}{$15.9$} & Gradients & $206$ & $5.9$ \\
    \cline{5-7}
    % (cont.): rm,wer,mwbe,f50 <- hyp-align/fastconformer-stream-rnnt-buckeye-segA-5h-native-metrics.txt
     &  &  &  & Posteriors & $318$ & $0.4$ \\
    \hline\hline
    % (cont.): rm,wer,mwbe,f50 <- hyp-align/whisper-base-charlev-buckeye-segA-5h-grad-metrics.txt
    \multirow{9}{*}{AED} & \multirow{2}{*}{\makecell[l]{Whisper-\\base}} & \multirow{2}{*}{$85.5$} & \multirow{2}{*}{$18.0$} & Gradients & $85$ & $25.5$ \\
    \cline{5-7}
    % (cont.): rm,wer,mwbe,f50 <- hyp-align/whisper-crossattn-buckeye-segA-5h-metrics.txt
     &  &  &  & Cross-att. & $77$ & $35.8$ \\
    \cline{2-7}
    % (cont.): rm,wer,mwbe,f50 <- hyp-align/whisper-large-v3-charlev-buckeye-segA-5h-grad-metrics.txt
     & \multirow{2}{*}{\makecell[l]{Whisper-\\large-v3}} & \multirow{2}{*}{$88.1$} & \multirow{2}{*}{$15.3$} & Gradients & $66$ & $40.4$ \\
    \cline{5-7}
    % (cont.): rm,wer,mwbe,f50 <- hyp-align/whisper-large-v3-charlev-buckeye-segA-5h-native-metrics.txt
     &  &  &  & Cross-att. & $66$ & $40.4$ \\
    \cline{2-7}
    % (cont.): rm,wer,mwbe,f50 <- hyp-align/crisperwhisper-charlev-buckeye-segA-5h-grad-metrics.txt
     & \multirow{3}{*}{\makecell[l]{Crisper-\\Whisper}} & \multirow{2}{*}{$92.0$} & \multirow{2}{*}{$12.7$} & Gradients & $56$ & $33.8$ \\
    \cline{5-7}
    % (cont.): rm,wer,mwbe,f50 <- hyp-align/crisperwhisper-charlev-buckeye-segA-5h-native-metrics.txt
     &  &  &  & Cross-att. & $44$ & $59.6$ \\
    \cline{3-7}
    % (cont.): rm,wer,mwbe,f50 <- hyp-align/crisperwhisper-official-buckeye-segA-5h-metrics.txt
     &  & $92.6$ & $12.2$ & Official & $46$ & $43.5$ \\
    \cline{2-7}
    % (cont.): rm,wer,mwbe,f50 <- hyp-align/owls-1B-180K-charlev-buckeye-segA-5h-grad-metrics.txt
     & \multirow{2}{*}{OWLS-1B} & \multirow{2}{*}{$85.7$} & \multirow{2}{*}{$208.0$} & Gradients & $1024$ & $3.8$ \\
    \cline{5-7}
    % (cont.): rm,wer,mwbe,f50 <- hyp-align/owls-1B-180K-charlev-buckeye-segA-5h-native-metrics.txt
     &  &  &  & Cross-att. & $1312$ & $7.8$ \\
    \hline\hline
    % (cont.): rm,wer,mwbe,f50 <- hyp-align/voxtral-charlevlogmel-buckeye-segA-5h-grad-metrics.txt
    \multirow{6}{*}{\makecell[l]{Speech \\ LLM}} & \multirow{2}{*}{Voxtral} & \multirow{2}{*}{$88.3$} & \multirow{2}{*}{$15.3$} & Gradients & $87$ & $25.3$ \\
    \cline{5-7}
    % (cont.): rm,wer,mwbe,f50 <- hyp-align/voxtral-charlevlogmel-buckeye-segA-5h-native-metrics.txt
     &  &  &  & Self-att. & $67$ & $37.4$ \\
    \cline{2-7}
    % (cont.): rm,wer,mwbe,f50 <- hyp-align/phi4mm-charlev-spc-buckeye-segA-5h-grad-metrics.txt
     & \multirow{2}{*}{Phi-4-MM} & \multirow{2}{*}{$92.8$} & \multirow{2}{*}{$11.7$} & Gradients & $102$ & $19.0$ \\
    \cline{5-7}
    % (cont.): rm,wer,mwbe,f50 <- hyp-align/phi4mm-charlev-spc-buckeye-segA-5h-native-metrics.txt
     &  &  &  & Self-att. & $80$ & $16.0$ \\
    \cline{2-7}
    % (cont.): rm,wer,mwbe,f50 <- hyp-align/canary-qwen-charlev-spc-logmel-st15-buckeye-segA-5h-grad-metrics.txt
     & \multirow{2}{*}{Canary-Qwen} & \multirow{2}{*}{$93.1$} & \multirow{2}{*}{$11.4$} & Gradients & $95$ & $17.1$ \\
    \cline{5-7}
    % (cont.): rm,wer,mwbe,f50 <- hyp-align/canary-qwen-charlev-spc-logmel-st15-buckeye-segA-5h-native-metrics.txt
     &  &  &  & Self-att. & $212$ & $3.9$ \\
    \hline
  \end{tabular}
  \end{adjustbox}
\end{table}

% =====================================================================
% AUTO-GENERATED by scripts/render_tables.py -- DO NOT EDIT THIS FILE.
% Regenerate from (edit these, then re-run render_tables.py / sync_tables_preview.sh):
%   tables-data/boundary-offsets.data.json    -- the numbers (from the Sisyphus graph)
%   tables-spec/boundary-offsets.spec.json     -- columns, units, layout
%   tables-spec/boundary-offsets.caption.tex   -- caption text
% =====================================================================
% data source (Sisyphus recipe): i6_experiments/users/zeyer/experiments/exp2025_07_07_in_grads/jobs/grad_align_tables.py :: _boundary_offset_table
% per-row comments below map columns -> the registered output name the number comes from
\begin{table}[tb]
  \centering
  \caption{%
\textbf{Signed boundary offsets} on Buckeye, gradient alignment vs.~the model's alternative aligner.
WBE is the mean absolute word-boundary error,
start/end off.~the signed mean boundary offset (positive = late),
and width err.~the signed word-width error (0 = correct duration).
Parakeet RNN-T is the offline contrast to the streaming transducers.%
}

  \label{tab:boundary-offsets}
  \begin{adjustbox}{max width=\linewidth}
  \begin{tabular}{|l|l|l|r|r|r|r|}
    \hline
    \multicolumn{2}{|c|}{\textbf{Model}} & \multirow{2}{*}{\textbf{\makecell[l]{Align\\method}}} & \multirow{2}{*}{\textbf{\makecell{WBE \\ {\normalfont {[ms]}\,$\downarrow$}}}} & \multirow{2}{*}{\textbf{\makecell{Start off. \\ {\normalfont {[ms]}}}}} & \multirow{2}{*}{\textbf{\makecell{End off. \\ {\normalfont {[ms]}}}}} & \multirow{2}{*}{\textbf{\makecell{Width err. \\ {\normalfont {[ms]}}}}} \\
    \cline{1-2}
    \textbf{Type} & \textbf{Name} &  &  &  &  &  \\ \hline\hline
    % (cont.): wbe,start,end,width <- align/wav2vec2ctc-fproj_out-prefixfwd-buckeye-segA-5h-L2_grad-pertoken-asotTrue-bs-5-en0.5-sil2.0-wordtopo-wbe.txt
    \multirow{10}{*}{CTC} & \multirow{2}{*}{MMS-FA} & Gradients & $121$ & $-93$ & $-91$ & $+2$ \\
    \cline{3-7}
    % (cont.): wbe,start,end,width <- baseline-mms_fa-buckeye-segA-5h-wbe.txt
     &  & Posteriors & $46$ & $+47$ & $-14$ & $-60$ \\
    \cline{2-7}
    % (cont.): wbe,start,end,width <- align/phoneme-vitouphy-prefixfwd-buckeye-segA-5h-L2_grad-pertoken-g2pword-asotTrue-bs-5-en0.5-sil2.0-wordtopo-wbe.txt
     & \multirow{2}{*}{\makecell[l]{XLS-R \\ (Phoneme)}} & Gradients & $81$ & $-72$ & $-47$ & $+25$ \\
    \cline{3-7}
    % (cont.): wbe,start,end,width <- baseline-phoneme-fa-buckeye-segA-5h-word-wbe.txt
     &  & Posteriors & $44$ & $+46$ & $-14$ & $-60$ \\
    \cline{2-7}
    % (cont.): wbe,start,end,width <- align/parakeet-ctc-1.1b-prefixfwd-buckeye-segA-5h-L2_grad-pertoken-asotTrue-bs-5-en0.5-sil2.0-wordtopo-wbe.txt
     & \multirow{2}{*}{\makecell[l]{Parakeet\\CTC}} & Gradients & $117$ & $-33$ & $-54$ & $-21$ \\
    \cline{3-7}
    % (cont.): wbe,start,end,width <- baseline-parakeet-ctc-1.1b-buckeye-segA-5h-wbe.txt
     &  & Posteriors & $99$ & $+53$ & $-51$ & $-104$ \\
    \cline{2-7}
    % (cont.): wbe,start,end,width <- align/owsm-ctc-v4-1b-lyr6-buckeye-segA-5h-L2_grad-pertoken-asotTrue-bs-5-en0.5-sil2.0-wordtopo-wbe.txt
     & \multirow{2}{*}{\makecell[l]{OWSM-\\CTC}} & Gradients & $145$ & $-89$ & $-107$ & $-18$ \\
    \cline{3-7}
    % (cont.): wbe,start,end,width <- baseline-owsm-ctc-v4-1b-lyr6-buckeye-segA-5h-wbe.txt
     &  & Posteriors & $110$ & $+125$ & $-10$ & $-136$ \\
    \cline{2-7}
    % (cont.): wbe,start,end,width <- align/fastconformer-stream-ctc-buckeye-segA-5h-L2_grad-pertoken-asotTrue-bs-5-en0.5-sil2.0-wordtopo-wbe.txt
     & \multirow{2}{*}{\makecell[l]{FastConformer\\(streaming)}} & Gradients & $158$ & $-62$ & $-97$ & $-35$ \\
    \cline{3-7}
    % (cont.): wbe,start,end,width <- baseline-fastconformer-stream-ctc-buckeye-segA-5h-wbe.txt
     &  & Posteriors & $366$ & $+415$ & $+310$ & $-106$ \\
    \hline\hline
    % (cont.): wbe,start,end,width <- align/parakeet-rnnt-1.1b-logmel-buckeye-segA-5h-L2_grad-pertoken-asotTrue-bs-5-en0.5-sil2.0-wordtopo-wbe.txt
    \multirow{6}{*}{Transd.} & \multirow{2}{*}{\makecell[l]{Parakeet\\RNN-T}} & Gradients & $147$ & $-22$ & $-42$ & $-20$ \\
    \cline{3-7}
    % (cont.): wbe,start,end,width <- baseline-parakeet-rnnt-1.1b-native-viterbi-buckeye-segA-5h-wbe.txt
     &  & Posteriors & $93$ & $+42$ & $-70$ & $-113$ \\
    \cline{2-7}
    % (cont.): wbe,start,end,width <- align/emformer-rnnt-prefix-logmel-buckeye-segA-5h-L2_grad-pertoken-asotTrue-bs-5-en0.5-sil2.0-wordtopo-wbe.txt
     & \multirow{2}{*}{\makecell[l]{Emformer\\(streaming)}} & Gradients & $139$ & $+67$ & $+32$ & $-35$ \\
    \cline{3-7}
    % (cont.): wbe,start,end,width <- baseline-emformer-rnnt-native-viterbi-buckeye-segA-5h-wbe.txt
     &  & Posteriors & $399$ & $+484$ & $+308$ & $-176$ \\
    \cline{2-7}
    % (cont.): wbe,start,end,width <- align/fastconformer-stream-rnnt-buckeye-segA-5h-L2_grad-pertoken-asotTrue-bs-5-en0.5-sil2.0-wordtopo-wbe.txt
     & \multirow{2}{*}{\makecell[l]{FastConformer\\(streaming)}} & Gradients & $189$ & $-85$ & $-123$ & $-37$ \\
    \cline{3-7}
    % (cont.): wbe,start,end,width <- baseline-fastconformer-stream-rnnt-native-viterbi-buckeye-segA-5h-wbe.txt
     &  & Posteriors & $331$ & $+386$ & $+260$ & $-126$ \\
    \hline\hline
    % (cont.): wbe,start,end,width <- align/whisper-base-logmel-buckeye-segA-5h-L2_grad-pertoken-charlev-spc-asotTrue-bs-5-en0.5-sil2.0-wordtopo-wbe.txt
    \multirow{2}{*}{AED} & \multirow{2}{*}{\makecell[l]{Whisper-\\base}} & Gradients & $75$ & $+4$ & $-4$ & $-9$ \\
    \cline{3-7}
    % (cont.): wbe,start,end,width <- align/baseline-whisper-base-crossattn-auto-buckeye-segA-5h-asotTrue-bs-5-en0.5-sil2.0-wordtopo-wbe.txt
     &  & Cross-att. & $62$ & $-38$ & $-19$ & $+19$ \\
    \hline
  \end{tabular}
  \end{adjustbox}
\end{table}

% =====================================================================
% AUTO-GENERATED by scripts/render_tables.py -- DO NOT EDIT THIS FILE.
% Regenerate from (edit these, then re-run render_tables.py / sync_tables_preview.sh):
%   tables-data/char-vs-subword-family.data.json    -- the numbers (from the Sisyphus graph)
%   tables-spec/char-vs-subword-family.spec.json     -- columns, units, layout
%   tables-spec/char-vs-subword-family.caption.tex   -- caption text
% =====================================================================
% data source (Sisyphus recipe): i6_experiments/users/zeyer/experiments/exp2025_07_07_in_grads/jobs/grad_align_tables.py :: _char_subword_family_table
% per-row comments below map columns -> the registered output name the number comes from
\begin{table}[tb]
  \centering
  \caption{%
\textbf{Character vs.~subword targets}, one representative model per family, Buckeye.
Each model shows its gradient alignment and its model-native alternative.
The character target is a valid re-segmentation of the transcript,
one token per character with a word-boundary token between words.%
}

  \label{tab:char-vs-subword-family}
  \begin{adjustbox}{max width=\linewidth}
  \begin{tabular}{|l|l|l|r|r|r|r|}
    \hline
    \multicolumn{2}{|c|}{\textbf{Model}} & \multirow{2}{*}{\textbf{\makecell[l]{Align\\method}}} & \multicolumn{2}{c|}{\textbf{Char}} & \multicolumn{2}{c|}{\textbf{Subword}} \\
    \cline{1-2} \cline{4-5} \cline{6-7}
    \textbf{Type} & \textbf{Name} &  & \textbf{\makecell{WBE \\ {\normalfont {[ms]}\,$\downarrow$}}} & \textbf{\makecell{$\le$50ms \\ {\normalfont {[\%]}\,$\uparrow$}}} & \textbf{\makecell{WBE \\ {\normalfont {[ms]}\,$\downarrow$}}} & \textbf{\makecell{$\le$50ms \\ {\normalfont {[\%]}\,$\uparrow$}}} \\ \hline\hline
    % (cont.): c_wbe,c_a50 <- align/parakeet-ctc-1.1b-prefixfwd-char-abl-buckeye-segA-5h-L2_grad-pertoken-asotTrue-bs-5-en0.5-sil2.0-wordtopo-wbe.txt  |  s_wbe,s_a50 <- align/parakeet-ctc-1.1b-prefixfwd-abl-buckeye-segA-5h-L2_grad-pertoken-asotTrue-bs-5-en0.5-sil2.0-wordtopo-wbe.txt
    \multirow{2}{*}{CTC} & \multirow{2}{*}{Parakeet CTC} & Gradients & $4167$ & $2.8$ & $117$ & $37.1$ \\
    \cline{3-7}
    % (cont.): c_wbe,c_a50 <- baseline-parakeet-ctc-1.1b-char-buckeye-segA-5h-wbe.txt  |  s_wbe,s_a50 <- baseline-parakeet-ctc-1.1b-buckeye-segA-5h-wbe.txt
     &  & Posteriors & $173$ & $26.1$ & $99$ & $32.2$ \\
    \hline\hline
    % (cont.): c_wbe,c_a50 <- align/parakeet-rnnt-1.1b-logmel-char-abl-buckeye-segA-5h-L2_grad-pertoken-asotTrue-bs-5-en0.5-sil2.0-wordtopo-wbe.txt  |  s_wbe,s_a50 <- align/parakeet-rnnt-1.1b-logmel-abl-buckeye-segA-5h-L2_grad-pertoken-asotTrue-bs-5-en0.5-sil2.0-wordtopo-wbe.txt
    \multirow{2}{*}{Transd.} & \multirow{2}{*}{Parakeet RNN-T} & Gradients & $384$ & $22.1$ & $147$ & $29.1$ \\
    \cline{3-7}
    % (cont.): c_wbe,c_a50 <- baseline-parakeet-rnnt-1.1b-char-native-viterbi-buckeye-segA-5h-wbe.txt  |  s_wbe,s_a50 <- baseline-parakeet-rnnt-1.1b-native-viterbi-buckeye-segA-5h-wbe.txt
     &  & Posteriors & $1255$ & $14.8$ & $93$ & $35.4$ \\
    \hline\hline
    % (cont.): c_wbe,c_a50 <- align/whisper-large-v3-logmel-buckeye-segA-5h-L2_grad-pertoken-charlev-spc-asotTrue-bs-5-en0.5-sil2.0-wordtopo-wbe.txt  |  s_wbe,s_a50 <- align/whisper-large-v3-logmel-buckeye-segA-5h-L2_grad-pertoken-subword-asotTrue-bs-5-en0.5-sil2.0-wordtopo-wbe.txt
    \multirow{2}{*}{AED} & \multirow{2}{*}{Whisper-large-v3} & Gradients & $53$ & $66.2$ & $108$ & $41.2$ \\
    \cline{3-7}
    % (cont.): c_wbe,c_a50 <- align/baseline-whisper-large-v3-crossattn-charlev-buckeye-segA-5h-asotTrue-bs-5-en0.5-sil2.0-wordtopo-wbe.txt  |  s_wbe,s_a50 <- align/baseline-whisper-large-v3-crossattn-auto-buckeye-segA-5h-asotTrue-bs-5-en0.5-sil2.0-wordtopo-wbe.txt
     &  & Cross-att. & $57$ & $75.2$ & $49$ & $69.1$ \\
    \hline\hline
    % (cont.): c_wbe,c_a50 <- align/voxtral-charlevlogmel-buckeye-segA-5h-L2_grad-pertoken-asotTrue-bs-5-en0.5-sil2.0-wordtopo-wbe.txt  |  s_wbe,s_a50 <- align/voxtral-logmel-buckeye-segA-5h-L2_grad-pertoken-subword-asotTrue-bs-5-en0.5-sil2.0-wordtopo-wbe.txt
    \multirow{2}{*}{\makecell[l]{Speech \\ LLM}} & \multirow{2}{*}{Voxtral} & Gradients & $74$ & $51.0$ & $118$ & $38.0$ \\
    \cline{3-7}
    % (cont.): c_wbe,c_a50 <- align/baseline-voxtral-char-selfattn-buckeye-segA-5h-asotTrue-bs-5-en0.5-sil2.0-wordtopo-wbe.txt  |  s_wbe,s_a50 <- align/baseline-voxtral-selfattn-buckeye-segA-5h-asotTrue-bs-5-en0.5-sil2.0-wordtopo-wbe.txt
     &  & Self-att. & $129$ & $46.0$ & $52$ & $65.5$ \\
    \hline
  \end{tabular}
  \end{adjustbox}
\end{table}

% =====================================================================
% AUTO-GENERATED by scripts/render_tables.py -- DO NOT EDIT THIS FILE.
% Regenerate from (edit these, then re-run render_tables.py / sync_tables_preview.sh):
%   tables-data/grad-score-ablation.data.json    -- the numbers (from the Sisyphus graph)
%   tables-spec/grad-score-ablation.spec.json     -- columns, units, layout
%   tables-spec/grad-score-ablation.caption.tex   -- caption text
% =====================================================================
% data source (Sisyphus recipe): i6_experiments/users/zeyer/experiments/exp2025_07_07_in_grads/jobs/grad_align_tables.py :: _ablation_table
% per-row comments below map columns -> the registered output name the number comes from
\begin{table}[tb]
  \centering
  \caption{%
\textbf{Grad-score ablation} across families, Buckeye.
The feature-axis reduction (L0.5 / L1 / L2 / sum), and whether the gradient is multiplied by the input (the right group).
The signed sum columns need special handling, since their score can be negative:
normalized score, constant blank score, on plain CTC topology.%
}

  \label{tab:gradscore}
  \begin{adjustbox}{max width=\linewidth}
  \begin{tabular}{|l|l|r|r|r|r|r|r|}
    \hline
    \multicolumn{2}{|c|}{\textbf{Model}} & \multicolumn{6}{c|}{\textbf{WBE\,{\normalfont {[ms]}\,$\downarrow$}}} \\
    \cline{1-2} \cline{3-8}
    \multirow{2}{*}{\textbf{Type}} & \multirow{2}{*}{\textbf{Name}} & \multicolumn{4}{c|}{\textbf{$\nabla$ (gradient)}} & \multicolumn{2}{c|}{\textbf{$\nabla \times$ input}} \\
    \cline{3-6} \cline{7-8}
     &  & \textbf{L0.5} & \textbf{L1} & \textbf{L2} & \textbf{sum} & \textbf{L2} & \textbf{sum} \\ \hline\hline
    % (cont.): L0.5_grad <- align/wav2vec2ctc-fproj_out-prefixfwd-abl-buckeye-segA-5h-L0.5_grad-pertoken-asotTrue-bs-5-en0.5-sil2.0-wordtopo-wbe.txt  |  L1_grad <- align/wav2vec2ctc-fproj_out-prefixfwd-abl-buckeye-segA-5h-L1_grad-pertoken-asotTrue-bs-5-en0.5-sil2.0-wordtopo-wbe.txt  |  L2_grad <- align/wav2vec2ctc-fproj_out-prefixfwd-abl-buckeye-segA-5h-L2_grad-pertoken-asotTrue-bs-5-en0.5-sil2.0-wordtopo-wbe.txt  |  dot_grad <- align/wav2vec2ctc-fproj_out-prefixfwd-abl-buckeye-segA-5h-dot_grad-pertoken-nsabsmeanS-nse0.05-cs1e-05_None-asotTrue-bs-6-wbe.txt  |  L2_e_grad <- align/wav2vec2ctc-fproj_out-prefixfwd-abl-buckeye-segA-5h-L2_e_grad-pertoken-asotTrue-bs-5-en0.5-sil2.0-wordtopo-wbe.txt  |  dot_e_grad <- align/wav2vec2ctc-fproj_out-prefixfwd-abl-buckeye-segA-5h-dot_e_grad-pertoken-nsabsmeanS-nse0.05-cs1e-05_None-asotTrue-bs-6-wbe.txt
    \multirow{2}{*}{CTC} & MMS-FA & $121$ & $121$ & $121$ & $197$ & $120$ & $164$ \\
    \cline{2-8}
    % (cont.): L0.5_grad <- align/parakeet-ctc-1.1b-prefixfwd-abl-buckeye-segA-5h-L0.5_grad-pertoken-asotTrue-bs-5-en0.5-sil2.0-wordtopo-wbe.txt  |  L1_grad <- align/parakeet-ctc-1.1b-prefixfwd-abl-buckeye-segA-5h-L1_grad-pertoken-asotTrue-bs-5-en0.5-sil2.0-wordtopo-wbe.txt  |  L2_grad <- align/parakeet-ctc-1.1b-prefixfwd-abl-buckeye-segA-5h-L2_grad-pertoken-asotTrue-bs-5-en0.5-sil2.0-wordtopo-wbe.txt  |  dot_grad <- align/parakeet-ctc-1.1b-prefixfwd-abl-buckeye-segA-5h-dot_grad-pertoken-nsabsmeanS-nse0.05-cs1e-05_None-asotTrue-bs-6-wbe.txt  |  L2_e_grad <- align/parakeet-ctc-1.1b-prefixfwd-abl-buckeye-segA-5h-L2_e_grad-pertoken-asotTrue-bs-5-en0.5-sil2.0-wordtopo-wbe.txt  |  dot_e_grad <- align/parakeet-ctc-1.1b-prefixfwd-abl-buckeye-segA-5h-dot_e_grad-pertoken-nsabsmeanS-nse0.05-cs1e-05_None-asotTrue-bs-6-wbe.txt
     & Parakeet CTC & $118$ & $117$ & $117$ & $172$ & $116$ & $166$ \\
    \hline
    % (cont.): L0.5_grad <- align/parakeet-rnnt-1.1b-logmel-abl-buckeye-segA-5h-L0.5_grad-pertoken-asotTrue-bs-5-en0.5-sil2.0-wordtopo-wbe.txt  |  L1_grad <- align/parakeet-rnnt-1.1b-logmel-abl-buckeye-segA-5h-L1_grad-pertoken-asotTrue-bs-5-en0.5-sil2.0-wordtopo-wbe.txt  |  L2_grad <- align/parakeet-rnnt-1.1b-logmel-abl-buckeye-segA-5h-L2_grad-pertoken-asotTrue-bs-5-en0.5-sil2.0-wordtopo-wbe.txt  |  dot_grad <- align/parakeet-rnnt-1.1b-logmel-abl-buckeye-segA-5h-dot_grad-pertoken-nsabsmeanS-nse0.05-cs1e-05_None-asotTrue-bs-6-wbe.txt  |  L2_e_grad <- align/parakeet-rnnt-1.1b-logmel-abl-buckeye-segA-5h-L2_e_grad-pertoken-asotTrue-bs-5-en0.5-sil2.0-wordtopo-wbe.txt  |  dot_e_grad <- align/parakeet-rnnt-1.1b-logmel-abl-buckeye-segA-5h-dot_e_grad-pertoken-nsabsmeanS-nse0.05-cs1e-05_None-asotTrue-bs-6-wbe.txt
    \multirow{2}{*}{Transd.} & Parakeet RNN-T & $147$ & $147$ & $147$ & $181$ & $146$ & $188$ \\
    \cline{2-8}
    % (cont.): L0.5_grad <- align/emformer-rnnt-prefix-logmel-abl-buckeye-segA-5h-L0.5_grad-pertoken-asotTrue-bs-5-en0.5-sil2.0-wordtopo-wbe.txt  |  L1_grad <- align/emformer-rnnt-prefix-logmel-abl-buckeye-segA-5h-L1_grad-pertoken-asotTrue-bs-5-en0.5-sil2.0-wordtopo-wbe.txt  |  L2_grad <- align/emformer-rnnt-prefix-logmel-abl-buckeye-segA-5h-L2_grad-pertoken-asotTrue-bs-5-en0.5-sil2.0-wordtopo-wbe.txt  |  dot_grad <- align/emformer-rnnt-prefix-logmel-abl-buckeye-segA-5h-dot_grad-pertoken-nsabsmeanS-nse0.05-cs1e-05_None-asotTrue-bs-6-wbe.txt  |  L2_e_grad <- align/emformer-rnnt-prefix-logmel-abl-buckeye-segA-5h-L2_e_grad-pertoken-asotTrue-bs-5-en0.5-sil2.0-wordtopo-wbe.txt  |  dot_e_grad <- align/emformer-rnnt-prefix-logmel-abl-buckeye-segA-5h-dot_e_grad-pertoken-nsabsmeanS-nse0.05-cs1e-05_None-asotTrue-bs-6-wbe.txt
     & \makecell[l]{Emformer\\(streaming)} & $142$ & $140$ & $139$ & $184$ & $138$ & $190$ \\
    \hline
    % (cont.): L0.5_grad <- align/whisper-base-logmel-charlev-spc-abl-buckeye-segA-5h-L0.5_grad-pertoken-asotTrue-bs-5-en0.5-sil2.0-wordtopo-wbe.txt  |  L1_grad <- align/whisper-base-logmel-charlev-spc-abl-buckeye-segA-5h-L1_grad-pertoken-asotTrue-bs-5-en0.5-sil2.0-wordtopo-wbe.txt  |  L2_grad <- align/whisper-base-logmel-charlev-spc-abl-buckeye-segA-5h-L2_grad-pertoken-asotTrue-bs-5-en0.5-sil2.0-wordtopo-wbe.txt  |  dot_grad <- align/whisper-base-logmel-charlev-spc-abl-buckeye-segA-5h-dot_grad-pertoken-nsabsmeanS-nse0.05-cs1e-05_None-asotTrue-bs-6-wbe.txt  |  L2_e_grad <- align/whisper-base-logmel-charlev-spc-abl-buckeye-segA-5h-L2_e_grad-pertoken-asotTrue-bs-5-en0.5-sil2.0-wordtopo-wbe.txt  |  dot_e_grad <- align/whisper-base-logmel-charlev-spc-abl-buckeye-segA-5h-dot_e_grad-pertoken-nsabsmeanS-nse0.05-cs1e-05_None-asotTrue-bs-6-wbe.txt
    AED & Whisper-base & $74$ & $74$ & $75$ & $110$ & $76$ & $122$ \\
    \hline
    % (cont.): L0.5_grad <- align/voxtral-charlevlogmel-abl-buckeye-segA-5h-L0.5_grad-pertoken-asotTrue-bs-5-en0.5-sil2.0-wordtopo-wbe.txt  |  L1_grad <- align/voxtral-charlevlogmel-abl-buckeye-segA-5h-L1_grad-pertoken-asotTrue-bs-5-en0.5-sil2.0-wordtopo-wbe.txt  |  L2_grad <- align/voxtral-charlevlogmel-abl-buckeye-segA-5h-L2_grad-pertoken-asotTrue-bs-5-en0.5-sil2.0-wordtopo-wbe.txt  |  dot_grad <- align/voxtral-charlevlogmel-abl-buckeye-segA-5h-dot_grad-pertoken-nsabsmeanS-nse0.05-cs1e-05_None-asotTrue-bs-6-wbe.txt  |  L2_e_grad <- align/voxtral-charlevlogmel-abl-buckeye-segA-5h-L2_e_grad-pertoken-asotTrue-bs-5-en0.5-sil2.0-wordtopo-wbe.txt  |  dot_e_grad <- align/voxtral-charlevlogmel-abl-buckeye-segA-5h-dot_e_grad-pertoken-nsabsmeanS-nse0.05-cs1e-05_None-asotTrue-bs-6-wbe.txt
    \multirow{3}{*}{\makecell[l]{Speech \\ LLM}} & Voxtral & $73$ & $74$ & $74$ & $112$ & $76$ & $121$ \\
    \cline{2-8}
    % (cont.): L0.5_grad <- align/phi4mm-charlev-spc-abl-buckeye-segA-5h-L0.5_grad-pertoken-asotTrue-bs-5-en0.5-sil2.0-wordtopo-wbe.txt  |  L1_grad <- align/phi4mm-charlev-spc-abl-buckeye-segA-5h-L1_grad-pertoken-asotTrue-bs-5-en0.5-sil2.0-wordtopo-wbe.txt  |  L2_grad <- align/phi4mm-charlev-spc-abl-buckeye-segA-5h-L2_grad-pertoken-asotTrue-bs-5-en0.5-sil2.0-wordtopo-wbe.txt  |  dot_grad <- align/phi4mm-charlev-spc-abl-buckeye-segA-5h-dot_grad-pertoken-nsabsmeanS-nse0.05-cs1e-05_None-asotTrue-bs-6-wbe.txt  |  L2_e_grad <- align/phi4mm-charlev-spc-abl-buckeye-segA-5h-L2_e_grad-pertoken-asotTrue-bs-5-en0.5-sil2.0-wordtopo-wbe.txt  |  dot_e_grad <- align/phi4mm-charlev-spc-abl-buckeye-segA-5h-dot_e_grad-pertoken-nsabsmeanS-nse0.05-cs1e-05_None-asotTrue-bs-6-wbe.txt
     & Phi-4-MM & $107$ & $106$ & $104$ & $138$ & $103$ & $119$ \\
    \cline{2-8}
    % (cont.): L0.5_grad <- align/canary-qwen-charlev-spc-logmel-st15-abl-buckeye-segA-5h-L0.5_grad-pertoken-asotTrue-bs-5-en0.5-sil2.0-wordtopo-wbe.txt  |  L1_grad <- align/canary-qwen-charlev-spc-logmel-st15-abl-buckeye-segA-5h-L1_grad-pertoken-asotTrue-bs-5-en0.5-sil2.0-wordtopo-wbe.txt  |  L2_grad <- align/canary-qwen-charlev-spc-logmel-st15-abl-buckeye-segA-5h-L2_grad-pertoken-asotTrue-bs-5-en0.5-sil2.0-wordtopo-wbe.txt  |  dot_grad <- align/canary-qwen-charlev-spc-logmel-st15-abl-buckeye-segA-5h-dot_grad-pertoken-nsabsmeanS-nse0.05-cs1e-05_None-asotTrue-bs-6-wbe.txt  |  L2_e_grad <- align/canary-qwen-charlev-spc-logmel-st15-abl-buckeye-segA-5h-L2_e_grad-pertoken-asotTrue-bs-5-en0.5-sil2.0-wordtopo-wbe.txt  |  dot_e_grad <- align/canary-qwen-charlev-spc-logmel-st15-abl-buckeye-segA-5h-dot_e_grad-pertoken-nsabsmeanS-nse0.05-cs1e-05_None-asotTrue-bs-6-wbe.txt
     & Canary-Qwen & $100$ & $100$ & $99$ & $107$ & $99$ & $113$ \\
    \hline
  \end{tabular}
  \end{adjustbox}
\end{table}

% =====================================================================
% AUTO-GENERATED by scripts/render_tables.py -- DO NOT EDIT THIS FILE.
% Regenerate from (edit these, then re-run render_tables.py / sync_tables_preview.sh):
%   tables-data/alignopts-silence.data.json    -- the numbers (from the Sisyphus graph)
%   tables-spec/alignopts-silence.spec.json     -- columns, units, layout
%   tables-spec/alignopts-silence.caption.tex   -- caption text
% =====================================================================
% data source (Sisyphus recipe): i6_experiments/users/zeyer/experiments/exp2025_07_07_in_grads/jobs/grad_align_tables.py :: _alignopts_silence_table
% per-row comments below map columns -> the registered output name the number comes from
\begin{table}[tb]
  \centering
  \caption{%
\textbf{Blank-scoring scheme} across signals, Buckeye, word-level topology, energy weighting fixed at $\rho{=}0.5$.
The constant blank $\gamma$, the energy-aware silence $\beta_t = \mu_t - \lambda\,z(E_t)\,\sigma_t$,
and the z-score $\beta_t = \mu_t + \kappa\,\sigma_t$.
WBE for the gradient signal and for the native attention signal.%
}

  \label{tab:alignopts-silence}
  \begin{adjustbox}{max width=\linewidth}
  \begin{tabular}{|c|c|r|r|r|r|r|r|r|}
    \hline
    \multirow{3}{*}{\textbf{Blank}} & \multirow{3}{*}{\textbf{Opts}} & \multicolumn{7}{c|}{\textbf{WBE\,{\normalfont {[ms]}\,$\downarrow$}}} \\
    \cline{3-9}
     &  & \multicolumn{4}{c|}{\textbf{Gradients}} & \multicolumn{2}{c|}{\textbf{Cross-att.}} & \multicolumn{1}{c|}{\textbf{Self-att.}} \\
    \cline{3-6} \cline{7-8} \cline{9-9}
     &  & \textbf{\makecell{MMS-\\FA}} & \textbf{\makecell{Whi-\\sper}} & \textbf{OWLS} & \textbf{\makecell{Vox-\\tral}} & \textbf{\makecell{Whi-\\sper}} & \textbf{OWLS} & \textbf{\makecell{Vox-\\tral}} \\ \hline\hline
    % (cont.): mms-fa <- align/wav2vec2ctc-fproj_out-prefixfwd-abl-buckeye-segA-5h-L2_grad-pertoken-asotTrue-bs-3-en0.5-wordtopo-wbe.txt  |  whisper <- align/whisper-large-v3-logmel-charlev-spc-abl-buckeye-segA-5h-L2_grad-pertoken-asotTrue-bs-3-en0.5-wordtopo-wbe.txt  |  owls <- align/owls-1B-180K-charlev-logmel-buckeye-segA-5h-L2_grad-pertoken-asotTrue-bs-3-en0.5-wordtopo-wbe.txt  |  voxtral <- align/voxtral-charlevlogmel-buckeye-segA-5h-L2_grad-pertoken-asotTrue-bs-3-en0.5-wordtopo-wbe.txt  |  wh_a <- align/baseline-whisper-large-v3-crossattn-auto-buckeye-segA-5h-asotTrue-bs-3-en0.5-wordtopo-wbe.txt  |  owls_a <- align/baseline-owls-1B-180K-crossattn-auto-buckeye-segA-5h-asotTrue-bs-3-en0.5-wordtopo-wbe.txt  |  vx_a <- align/baseline-voxtral-selfattn-buckeye-segA-5h-asotTrue-bs-3-en0.5-wordtopo-wbe.txt
    \multirow{3}{*}{constant} & $\gamma = -3$ & $176$ & $129$ & $658$ & $183$ & $100$ & $74$ & $63$ \\
    \cline{2-9}
    % (cont.): mms-fa <- align/wav2vec2ctc-fproj_out-prefixfwd-abl-buckeye-segA-5h-L2_grad-pertoken-asotTrue-bs-5-en0.5-wordtopo-wbe.txt  |  whisper <- align/whisper-large-v3-logmel-charlev-spc-abl-buckeye-segA-5h-L2_grad-pertoken-asotTrue-bs-5-en0.5-wordtopo-wbe.txt  |  owls <- align/owls-1B-180K-charlev-logmel-buckeye-segA-5h-L2_grad-pertoken-asotTrue-bs-5-en0.5-wordtopo-wbe.txt  |  voxtral <- align/voxtral-charlevlogmel-buckeye-segA-5h-L2_grad-pertoken-asotTrue-bs-5-en0.5-wordtopo-wbe.txt  |  wh_a <- align/baseline-whisper-large-v3-crossattn-auto-buckeye-segA-5h-asotTrue-bs-5-en0.5-wordtopo-wbe.txt  |  owls_a <- align/baseline-owls-1B-180K-crossattn-auto-buckeye-segA-5h-asotTrue-bs-5-en0.5-wordtopo-wbe.txt  |  vx_a <- align/baseline-voxtral-selfattn-buckeye-segA-5h-asotTrue-bs-5-en0.5-wordtopo-wbe.txt
     & $\gamma = -5$ & $131$ & $87$ & $577$ & $131$ & $44$ & $65$ & $53$ \\
    \cline{2-9}
    % (cont.): mms-fa <- align/wav2vec2ctc-fproj_out-prefixfwd-abl-buckeye-segA-5h-L2_grad-pertoken-asotTrue-bs-8-en0.5-wordtopo-wbe.txt  |  whisper <- align/whisper-large-v3-logmel-charlev-spc-abl-buckeye-segA-5h-L2_grad-pertoken-asotTrue-bs-8-en0.5-wordtopo-wbe.txt  |  owls <- align/owls-1B-180K-charlev-logmel-buckeye-segA-5h-L2_grad-pertoken-asotTrue-bs-8-en0.5-wordtopo-wbe.txt  |  voxtral <- align/voxtral-charlevlogmel-buckeye-segA-5h-L2_grad-pertoken-asotTrue-bs-8-en0.5-wordtopo-wbe.txt  |  wh_a <- align/baseline-whisper-large-v3-crossattn-auto-buckeye-segA-5h-asotTrue-bs-8-en0.5-wordtopo-wbe.txt  |  owls_a <- align/baseline-owls-1B-180K-crossattn-auto-buckeye-segA-5h-asotTrue-bs-8-en0.5-wordtopo-wbe.txt  |  vx_a <- align/baseline-voxtral-selfattn-buckeye-segA-5h-asotTrue-bs-8-en0.5-wordtopo-wbe.txt
     & $\gamma = -8$ & $141$ & $65$ & $237$ & $94$ & $65$ & $70$ & $65$ \\
    \hline
    % (cont.): mms-fa <- align/wav2vec2ctc-fproj_out-prefixfwd-abl-buckeye-segA-5h-L2_grad-pertoken-asotTrue-bs-5-en0.5-sil0.5-wordtopo-wbe.txt  |  whisper <- align/whisper-large-v3-logmel-charlev-spc-abl-buckeye-segA-5h-L2_grad-pertoken-asotTrue-bs-5-en0.5-sil0.5-wordtopo-wbe.txt  |  owls <- align/owls-1B-180K-charlev-logmel-buckeye-segA-5h-L2_grad-pertoken-asotTrue-bs-5-en0.5-sil0.5-wordtopo-wbe.txt  |  voxtral <- align/voxtral-charlevlogmel-buckeye-segA-5h-L2_grad-pertoken-asotTrue-bs-5-en0.5-sil0.5-wordtopo-wbe.txt  |  wh_a <- align/baseline-whisper-large-v3-crossattn-auto-buckeye-segA-5h-asotTrue-bs-5-en0.5-sil0.5-wordtopo-wbe.txt  |  owls_a <- align/baseline-owls-1B-180K-crossattn-auto-buckeye-segA-5h-asotTrue-bs-5-en0.5-sil0.5-wordtopo-wbe.txt  |  vx_a <- align/baseline-voxtral-selfattn-buckeye-segA-5h-asotTrue-bs-5-en0.5-sil0.5-wordtopo-wbe.txt
    \multirow{4}{*}{energy} & $\lambda = 0.5$ & $131$ & $54$ & $193$ & $77$ & $63$ & $65$ & $59$ \\
    \cline{2-9}
    % (cont.): mms-fa <- align/wav2vec2ctc-fproj_out-prefixfwd-abl-buckeye-segA-5h-L2_grad-pertoken-asotTrue-bs-5-en0.5-sil1.0-wordtopo-wbe.txt  |  whisper <- align/whisper-large-v3-logmel-charlev-spc-abl-buckeye-segA-5h-L2_grad-pertoken-asotTrue-bs-5-en0.5-sil1.0-wordtopo-wbe.txt  |  owls <- align/owls-1B-180K-charlev-logmel-buckeye-segA-5h-L2_grad-pertoken-asotTrue-bs-5-en0.5-sil1.0-wordtopo-wbe.txt  |  voxtral <- align/voxtral-charlevlogmel-buckeye-segA-5h-L2_grad-pertoken-asotTrue-bs-5-en0.5-sil1.0-wordtopo-wbe.txt  |  wh_a <- align/baseline-whisper-large-v3-crossattn-auto-buckeye-segA-5h-asotTrue-bs-5-en0.5-sil1.0-wordtopo-wbe.txt  |  owls_a <- align/baseline-owls-1B-180K-crossattn-auto-buckeye-segA-5h-asotTrue-bs-5-en0.5-sil1.0-wordtopo-wbe.txt  |  vx_a <- align/baseline-voxtral-selfattn-buckeye-segA-5h-asotTrue-bs-5-en0.5-sil1.0-wordtopo-wbe.txt
     & $\lambda = 1$ & $127$ & $53$ & $188$ & $75$ & $61$ & $60$ & $56$ \\
    \cline{2-9}
    % (cont.): mms-fa <- align/wav2vec2ctc-fproj_out-prefixfwd-abl-buckeye-segA-5h-L2_grad-pertoken-asotTrue-bs-5-en0.5-sil2.0-wordtopo-wbe.txt  |  whisper <- align/whisper-large-v3-logmel-charlev-spc-abl-buckeye-segA-5h-L2_grad-pertoken-asotTrue-bs-5-en0.5-sil2.0-wordtopo-wbe.txt  |  owls <- align/owls-1B-180K-charlev-logmel-buckeye-segA-5h-L2_grad-pertoken-asotTrue-bs-5-en0.5-sil2.0-wordtopo-wbe.txt  |  voxtral <- align/voxtral-charlevlogmel-buckeye-segA-5h-L2_grad-pertoken-asotTrue-bs-5-en0.5-sil2.0-wordtopo-wbe.txt  |  wh_a <- align/baseline-whisper-large-v3-crossattn-auto-buckeye-segA-5h-asotTrue-bs-5-en0.5-sil2.0-wordtopo-wbe.txt  |  owls_a <- align/baseline-owls-1B-180K-crossattn-auto-buckeye-segA-5h-asotTrue-bs-5-en0.5-sil2.0-wordtopo-wbe.txt  |  vx_a <- align/baseline-voxtral-selfattn-buckeye-segA-5h-asotTrue-bs-5-en0.5-sil2.0-wordtopo-wbe.txt
     & $\lambda = 2$ & $121$ & $53$ & $180$ & $74$ & $49$ & $51$ & $52$ \\
    \cline{2-9}
    % (cont.): mms-fa <- align/wav2vec2ctc-fproj_out-prefixfwd-abl-buckeye-segA-5h-L2_grad-pertoken-asotTrue-bs-5-en0.5-sil3.0-wordtopo-wbe.txt  |  whisper <- align/whisper-large-v3-logmel-charlev-spc-abl-buckeye-segA-5h-L2_grad-pertoken-asotTrue-bs-5-en0.5-sil3.0-wordtopo-wbe.txt  |  owls <- align/owls-1B-180K-charlev-logmel-buckeye-segA-5h-L2_grad-pertoken-asotTrue-bs-5-en0.5-sil3.0-wordtopo-wbe.txt  |  voxtral <- align/voxtral-charlevlogmel-buckeye-segA-5h-L2_grad-pertoken-asotTrue-bs-5-en0.5-sil3.0-wordtopo-wbe.txt  |  wh_a <- align/baseline-whisper-large-v3-crossattn-auto-buckeye-segA-5h-asotTrue-bs-5-en0.5-sil3.0-wordtopo-wbe.txt  |  owls_a <- align/baseline-owls-1B-180K-crossattn-auto-buckeye-segA-5h-asotTrue-bs-5-en0.5-sil3.0-wordtopo-wbe.txt  |  vx_a <- align/baseline-voxtral-selfattn-buckeye-segA-5h-asotTrue-bs-5-en0.5-sil3.0-wordtopo-wbe.txt
     & $\lambda = 3$ & $117$ & $54$ & $174$ & $75$ & $48$ & $48$ & $55$ \\
    \hline
    % (cont.): mms-fa <- align/wav2vec2ctc-fproj_out-prefixfwd-abl-buckeye-segA-5h-L2_grad-pertoken-asotTrue-bs-5-en0.5-zsk0.5-wordtopo-wbe.txt  |  whisper <- align/whisper-large-v3-logmel-charlev-spc-abl-buckeye-segA-5h-L2_grad-pertoken-asotTrue-bs-5-en0.5-zsk0.5-wordtopo-wbe.txt  |  owls <- align/owls-1B-180K-charlev-logmel-buckeye-segA-5h-L2_grad-pertoken-asotTrue-bs-5-en0.5-zsk0.5-wordtopo-wbe.txt  |  voxtral <- align/voxtral-charlevlogmel-buckeye-segA-5h-L2_grad-pertoken-asotTrue-bs-5-en0.5-zsk0.5-wordtopo-wbe.txt  |  wh_a <- align/baseline-whisper-large-v3-crossattn-auto-buckeye-segA-5h-asotTrue-bs-5-en0.5-zsk0.5-wordtopo-wbe.txt  |  owls_a <- align/baseline-owls-1B-180K-crossattn-auto-buckeye-segA-5h-asotTrue-bs-5-en0.5-zsk0.5-wordtopo-wbe.txt  |  vx_a <- align/baseline-voxtral-selfattn-buckeye-segA-5h-asotTrue-bs-5-en0.5-zsk0.5-wordtopo-wbe.txt
    \multirow{3}{*}{z-score} & $\kappa = 0.5$ & $130$ & $54$ & $192$ & $77$ & $63$ & $65$ & $59$ \\
    \cline{2-9}
    % (cont.): mms-fa <- align/wav2vec2ctc-fproj_out-prefixfwd-abl-buckeye-segA-5h-L2_grad-pertoken-asotTrue-bs-5-en0.5-zsk1.0-wordtopo-wbe.txt  |  whisper <- align/whisper-large-v3-logmel-charlev-spc-abl-buckeye-segA-5h-L2_grad-pertoken-asotTrue-bs-5-en0.5-zsk1.0-wordtopo-wbe.txt  |  owls <- align/owls-1B-180K-charlev-logmel-buckeye-segA-5h-L2_grad-pertoken-asotTrue-bs-5-en0.5-zsk1.0-wordtopo-wbe.txt  |  voxtral <- align/voxtral-charlevlogmel-buckeye-segA-5h-L2_grad-pertoken-asotTrue-bs-5-en0.5-zsk1.0-wordtopo-wbe.txt  |  wh_a <- align/baseline-whisper-large-v3-crossattn-auto-buckeye-segA-5h-asotTrue-bs-5-en0.5-zsk1.0-wordtopo-wbe.txt  |  owls_a <- align/baseline-owls-1B-180K-crossattn-auto-buckeye-segA-5h-asotTrue-bs-5-en0.5-zsk1.0-wordtopo-wbe.txt  |  vx_a <- align/baseline-voxtral-selfattn-buckeye-segA-5h-asotTrue-bs-5-en0.5-zsk1.0-wordtopo-wbe.txt
     & $\kappa = 1$ & $127$ & $53$ & $186$ & $75$ & $61$ & $59$ & $56$ \\
    \cline{2-9}
    % (cont.): mms-fa <- align/wav2vec2ctc-fproj_out-prefixfwd-abl-buckeye-segA-5h-L2_grad-pertoken-asotTrue-bs-5-en0.5-zsk2.0-wordtopo-wbe.txt  |  whisper <- align/whisper-large-v3-logmel-charlev-spc-abl-buckeye-segA-5h-L2_grad-pertoken-asotTrue-bs-5-en0.5-zsk2.0-wordtopo-wbe.txt  |  owls <- align/owls-1B-180K-charlev-logmel-buckeye-segA-5h-L2_grad-pertoken-asotTrue-bs-5-en0.5-zsk2.0-wordtopo-wbe.txt  |  voxtral <- align/voxtral-charlevlogmel-buckeye-segA-5h-L2_grad-pertoken-asotTrue-bs-5-en0.5-zsk2.0-wordtopo-wbe.txt  |  wh_a <- align/baseline-whisper-large-v3-crossattn-auto-buckeye-segA-5h-asotTrue-bs-5-en0.5-zsk2.0-wordtopo-wbe.txt  |  owls_a <- align/baseline-owls-1B-180K-crossattn-auto-buckeye-segA-5h-asotTrue-bs-5-en0.5-zsk2.0-wordtopo-wbe.txt  |  vx_a <- align/baseline-voxtral-selfattn-buckeye-segA-5h-asotTrue-bs-5-en0.5-zsk2.0-wordtopo-wbe.txt
     & $\kappa = 2$ & $123$ & $56$ & $181$ & $78$ & $54$ & $51$ & $60$ \\
    \hline
  \end{tabular}
  \end{adjustbox}
\end{table}

% =====================================================================
% AUTO-GENERATED by scripts/render_tables.py -- DO NOT EDIT THIS FILE.
% Regenerate from (edit these, then re-run render_tables.py / sync_tables_preview.sh):
%   tables-data/alignopts-energy.data.json    -- the numbers (from the Sisyphus graph)
%   tables-spec/alignopts-energy.spec.json     -- columns, units, layout
%   tables-spec/alignopts-energy.caption.tex   -- caption text
% =====================================================================
% data source (Sisyphus recipe): i6_experiments/users/zeyer/experiments/exp2025_07_07_in_grads/jobs/grad_align_tables.py :: _alignopts_energy_table
% per-row comments below map columns -> the registered output name the number comes from
\begin{table}[tb]
  \centering
  \caption{%
\textbf{Audio-energy token weighting}, Buckeye, word-level topology.
The token scores are weighted by (audio energy)$^{\rho}$ before the DP,
and we sweep $\rho$ ($\rho{=}0$ disables it) at each blank scheme.
This complements \cref{tab:alignopts-silence}, which sweeps the blank scheme at fixed $\rho{=}0.5$.%
}

  \label{tab:alignopts-energy}
  \begin{adjustbox}{max width=\linewidth}
  \begin{tabular}{|c|c|r|r|r|r|r|r|r|}
    \hline
    \multirow{3}{*}{\textbf{Blank}} & \multirow{3}{*}{\textbf{$\rho$}} & \multicolumn{7}{c|}{\textbf{WBE\,{\normalfont {[ms]}\,$\downarrow$}}} \\
    \cline{3-9}
     &  & \multicolumn{4}{c|}{\textbf{Gradients}} & \multicolumn{2}{c|}{\textbf{Cross-att.}} & \multicolumn{1}{c|}{\textbf{Self-att.}} \\
    \cline{3-6} \cline{7-8} \cline{9-9}
     &  & \textbf{\makecell{MMS-\\FA}} & \textbf{\makecell{Whi-\\sper}} & \textbf{OWLS} & \textbf{\makecell{Vox-\\tral}} & \textbf{\makecell{Whi-\\sper}} & \textbf{OWLS} & \textbf{\makecell{Vox-\\tral}} \\ \hline\hline
    % (cont.): mms-fa <- align/wav2vec2ctc-fproj_out-prefixfwd-abl-buckeye-segA-5h-L2_grad-pertoken-asotTrue-bs-5-en0.0-wordtopo-wbe.txt  |  whisper <- align/whisper-large-v3-logmel-charlev-spc-abl-buckeye-segA-5h-L2_grad-pertoken-asotTrue-bs-5-en0.0-wordtopo-wbe.txt  |  owls <- align/owls-1B-180K-charlev-logmel-buckeye-segA-5h-L2_grad-pertoken-asotTrue-bs-5-en0.0-wordtopo-wbe.txt  |  voxtral <- align/voxtral-charlevlogmel-buckeye-segA-5h-L2_grad-pertoken-asotTrue-bs-5-en0.0-wordtopo-wbe.txt  |  wh_a <- align/baseline-whisper-large-v3-crossattn-auto-buckeye-segA-5h-asotTrue-bs-5-en0.0-wordtopo-wbe.txt  |  owls_a <- align/baseline-owls-1B-180K-crossattn-auto-buckeye-segA-5h-asotTrue-bs-5-en0.0-wordtopo-wbe.txt  |  vx_a <- align/baseline-voxtral-selfattn-buckeye-segA-5h-asotTrue-bs-5-en0.0-wordtopo-wbe.txt
    \multirow{5}{*}{\makecell[l]{constant \\ $\gamma = -5$}} & 0.0 & $147$ & $99$ & $884$ & $174$ & $54$ & $77$ & $59$ \\
    \cline{2-9}
    % (cont.): mms-fa <- align/wav2vec2ctc-fproj_out-prefixfwd-abl-buckeye-segA-5h-L2_grad-pertoken-asotTrue-bs-5-en0.25-wordtopo-wbe.txt  |  whisper <- align/whisper-large-v3-logmel-charlev-spc-abl-buckeye-segA-5h-L2_grad-pertoken-asotTrue-bs-5-en0.25-wordtopo-wbe.txt  |  owls <- align/owls-1B-180K-charlev-logmel-buckeye-segA-5h-L2_grad-pertoken-asotTrue-bs-5-en0.25-wordtopo-wbe.txt  |  voxtral <- align/voxtral-charlevlogmel-buckeye-segA-5h-L2_grad-pertoken-asotTrue-bs-5-en0.25-wordtopo-wbe.txt  |  wh_a <- align/baseline-whisper-large-v3-crossattn-auto-buckeye-segA-5h-asotTrue-bs-5-en0.25-wordtopo-wbe.txt  |  owls_a <- align/baseline-owls-1B-180K-crossattn-auto-buckeye-segA-5h-asotTrue-bs-5-en0.25-wordtopo-wbe.txt  |  vx_a <- align/baseline-voxtral-selfattn-buckeye-segA-5h-asotTrue-bs-5-en0.25-wordtopo-wbe.txt
     & 0.25 & $140$ & $89$ & $700$ & $141$ & $48$ & $71$ & $56$ \\
    \cline{2-9}
    % (cont.): mms-fa <- align/wav2vec2ctc-fproj_out-prefixfwd-abl-buckeye-segA-5h-L2_grad-pertoken-asotTrue-bs-5-en0.5-wordtopo-wbe.txt  |  whisper <- align/whisper-large-v3-logmel-charlev-spc-abl-buckeye-segA-5h-L2_grad-pertoken-asotTrue-bs-5-en0.5-wordtopo-wbe.txt  |  owls <- align/owls-1B-180K-charlev-logmel-buckeye-segA-5h-L2_grad-pertoken-asotTrue-bs-5-en0.5-wordtopo-wbe.txt  |  voxtral <- align/voxtral-charlevlogmel-buckeye-segA-5h-L2_grad-pertoken-asotTrue-bs-5-en0.5-wordtopo-wbe.txt  |  wh_a <- align/baseline-whisper-large-v3-crossattn-auto-buckeye-segA-5h-asotTrue-bs-5-en0.5-wordtopo-wbe.txt  |  owls_a <- align/baseline-owls-1B-180K-crossattn-auto-buckeye-segA-5h-asotTrue-bs-5-en0.5-wordtopo-wbe.txt  |  vx_a <- align/baseline-voxtral-selfattn-buckeye-segA-5h-asotTrue-bs-5-en0.5-wordtopo-wbe.txt
     & 0.5 & $131$ & $87$ & $577$ & $131$ & $44$ & $65$ & $53$ \\
    \cline{2-9}
    % (cont.): mms-fa <- align/wav2vec2ctc-fproj_out-prefixfwd-abl-buckeye-segA-5h-L2_grad-pertoken-asotTrue-bs-5-en0.75-wordtopo-wbe.txt  |  whisper <- align/whisper-large-v3-logmel-charlev-spc-abl-buckeye-segA-5h-L2_grad-pertoken-asotTrue-bs-5-en0.75-wordtopo-wbe.txt  |  owls <- align/owls-1B-180K-charlev-logmel-buckeye-segA-5h-L2_grad-pertoken-asotTrue-bs-5-en0.75-wordtopo-wbe.txt  |  voxtral <- align/voxtral-charlevlogmel-buckeye-segA-5h-L2_grad-pertoken-asotTrue-bs-5-en0.75-wordtopo-wbe.txt  |  wh_a <- align/baseline-whisper-large-v3-crossattn-auto-buckeye-segA-5h-asotTrue-bs-5-en0.75-wordtopo-wbe.txt  |  owls_a <- align/baseline-owls-1B-180K-crossattn-auto-buckeye-segA-5h-asotTrue-bs-5-en0.75-wordtopo-wbe.txt  |  vx_a <- align/baseline-voxtral-selfattn-buckeye-segA-5h-asotTrue-bs-5-en0.75-wordtopo-wbe.txt
     & 0.75 & $126$ & $89$ & $512$ & $131$ & $45$ & $60$ & $52$ \\
    \cline{2-9}
    % (cont.): mms-fa <- align/wav2vec2ctc-fproj_out-prefixfwd-abl-buckeye-segA-5h-L2_grad-pertoken-asotTrue-bs-5-en1.0-wordtopo-wbe.txt  |  whisper <- align/whisper-large-v3-logmel-charlev-spc-abl-buckeye-segA-5h-L2_grad-pertoken-asotTrue-bs-5-en1.0-wordtopo-wbe.txt  |  owls <- align/owls-1B-180K-charlev-logmel-buckeye-segA-5h-L2_grad-pertoken-asotTrue-bs-5-en1.0-wordtopo-wbe.txt  |  voxtral <- align/voxtral-charlevlogmel-buckeye-segA-5h-L2_grad-pertoken-asotTrue-bs-5-en1.0-wordtopo-wbe.txt  |  wh_a <- align/baseline-whisper-large-v3-crossattn-auto-buckeye-segA-5h-asotTrue-bs-5-en1.0-wordtopo-wbe.txt  |  owls_a <- align/baseline-owls-1B-180K-crossattn-auto-buckeye-segA-5h-asotTrue-bs-5-en1.0-wordtopo-wbe.txt  |  vx_a <- align/baseline-voxtral-selfattn-buckeye-segA-5h-asotTrue-bs-5-en1.0-wordtopo-wbe.txt
     & 1.0 & $124$ & $92$ & $485$ & $137$ & $47$ & $57$ & $52$ \\
    \hline\hline
    % (cont.): mms-fa <- align/wav2vec2ctc-fproj_out-prefixfwd-abl-buckeye-segA-5h-L2_grad-pertoken-asotTrue-bs-5-en0.0-sil1.0-wordtopo-wbe.txt  |  whisper <- align/whisper-large-v3-logmel-charlev-spc-abl-buckeye-segA-5h-L2_grad-pertoken-asotTrue-bs-5-en0.0-sil1.0-wordtopo-wbe.txt  |  owls <- align/owls-1B-180K-charlev-logmel-buckeye-segA-5h-L2_grad-pertoken-asotTrue-bs-5-en0.0-sil1.0-wordtopo-wbe.txt  |  voxtral <- align/voxtral-charlevlogmel-buckeye-segA-5h-L2_grad-pertoken-asotTrue-bs-5-en0.0-sil1.0-wordtopo-wbe.txt  |  wh_a <- align/baseline-whisper-large-v3-crossattn-auto-buckeye-segA-5h-asotTrue-bs-5-en0.0-sil1.0-wordtopo-wbe.txt  |  owls_a <- align/baseline-owls-1B-180K-crossattn-auto-buckeye-segA-5h-asotTrue-bs-5-en0.0-sil1.0-wordtopo-wbe.txt  |  vx_a <- align/baseline-voxtral-selfattn-buckeye-segA-5h-asotTrue-bs-5-en0.0-sil1.0-wordtopo-wbe.txt
    \multirow{5}{*}{\makecell[l]{energy \\ $\lambda = 1$}} & 0.0 & $141$ & $70$ & $263$ & $94$ & $64$ & $67$ & $62$ \\
    \cline{2-9}
    % (cont.): mms-fa <- align/wav2vec2ctc-fproj_out-prefixfwd-abl-buckeye-segA-5h-L2_grad-pertoken-asotTrue-bs-5-en0.25-sil1.0-wordtopo-wbe.txt  |  whisper <- align/whisper-large-v3-logmel-charlev-spc-abl-buckeye-segA-5h-L2_grad-pertoken-asotTrue-bs-5-en0.25-sil1.0-wordtopo-wbe.txt  |  owls <- align/owls-1B-180K-charlev-logmel-buckeye-segA-5h-L2_grad-pertoken-asotTrue-bs-5-en0.25-sil1.0-wordtopo-wbe.txt  |  voxtral <- align/voxtral-charlevlogmel-buckeye-segA-5h-L2_grad-pertoken-asotTrue-bs-5-en0.25-sil1.0-wordtopo-wbe.txt  |  wh_a <- align/baseline-whisper-large-v3-crossattn-auto-buckeye-segA-5h-asotTrue-bs-5-en0.25-sil1.0-wordtopo-wbe.txt  |  owls_a <- align/baseline-owls-1B-180K-crossattn-auto-buckeye-segA-5h-asotTrue-bs-5-en0.25-sil1.0-wordtopo-wbe.txt  |  vx_a <- align/baseline-voxtral-selfattn-buckeye-segA-5h-asotTrue-bs-5-en0.25-sil1.0-wordtopo-wbe.txt
     & 0.25 & $136$ & $58$ & $212$ & $81$ & $63$ & $65$ & $59$ \\
    \cline{2-9}
    % (cont.): mms-fa <- align/wav2vec2ctc-fproj_out-prefixfwd-abl-buckeye-segA-5h-L2_grad-pertoken-asotTrue-bs-5-en0.5-sil1.0-wordtopo-wbe.txt  |  whisper <- align/whisper-large-v3-logmel-charlev-spc-abl-buckeye-segA-5h-L2_grad-pertoken-asotTrue-bs-5-en0.5-sil1.0-wordtopo-wbe.txt  |  owls <- align/owls-1B-180K-charlev-logmel-buckeye-segA-5h-L2_grad-pertoken-asotTrue-bs-5-en0.5-sil1.0-wordtopo-wbe.txt  |  voxtral <- align/voxtral-charlevlogmel-buckeye-segA-5h-L2_grad-pertoken-asotTrue-bs-5-en0.5-sil1.0-wordtopo-wbe.txt  |  wh_a <- align/baseline-whisper-large-v3-crossattn-auto-buckeye-segA-5h-asotTrue-bs-5-en0.5-sil1.0-wordtopo-wbe.txt  |  owls_a <- align/baseline-owls-1B-180K-crossattn-auto-buckeye-segA-5h-asotTrue-bs-5-en0.5-sil1.0-wordtopo-wbe.txt  |  vx_a <- align/baseline-voxtral-selfattn-buckeye-segA-5h-asotTrue-bs-5-en0.5-sil1.0-wordtopo-wbe.txt
     & 0.5 & $127$ & $53$ & $188$ & $75$ & $61$ & $60$ & $56$ \\
    \cline{2-9}
    % (cont.): mms-fa <- align/wav2vec2ctc-fproj_out-prefixfwd-abl-buckeye-segA-5h-L2_grad-pertoken-asotTrue-bs-5-en0.75-sil1.0-wordtopo-wbe.txt  |  whisper <- align/whisper-large-v3-logmel-charlev-spc-abl-buckeye-segA-5h-L2_grad-pertoken-asotTrue-bs-5-en0.75-sil1.0-wordtopo-wbe.txt  |  owls <- align/owls-1B-180K-charlev-logmel-buckeye-segA-5h-L2_grad-pertoken-asotTrue-bs-5-en0.75-sil1.0-wordtopo-wbe.txt  |  voxtral <- align/voxtral-charlevlogmel-buckeye-segA-5h-L2_grad-pertoken-asotTrue-bs-5-en0.75-sil1.0-wordtopo-wbe.txt  |  wh_a <- align/baseline-whisper-large-v3-crossattn-auto-buckeye-segA-5h-asotTrue-bs-5-en0.75-sil1.0-wordtopo-wbe.txt  |  owls_a <- align/baseline-owls-1B-180K-crossattn-auto-buckeye-segA-5h-asotTrue-bs-5-en0.75-sil1.0-wordtopo-wbe.txt  |  vx_a <- align/baseline-voxtral-selfattn-buckeye-segA-5h-asotTrue-bs-5-en0.75-sil1.0-wordtopo-wbe.txt
     & 0.75 & $119$ & $53$ & $177$ & $75$ & $57$ & $57$ & $54$ \\
    \cline{2-9}
    % (cont.): mms-fa <- align/wav2vec2ctc-fproj_out-prefixfwd-abl-buckeye-segA-5h-L2_grad-pertoken-asotTrue-bs-5-en1.0-sil1.0-wordtopo-wbe.txt  |  whisper <- align/whisper-large-v3-logmel-charlev-spc-abl-buckeye-segA-5h-L2_grad-pertoken-asotTrue-bs-5-en1.0-sil1.0-wordtopo-wbe.txt  |  owls <- align/owls-1B-180K-charlev-logmel-buckeye-segA-5h-L2_grad-pertoken-asotTrue-bs-5-en1.0-sil1.0-wordtopo-wbe.txt  |  voxtral <- align/voxtral-charlevlogmel-buckeye-segA-5h-L2_grad-pertoken-asotTrue-bs-5-en1.0-sil1.0-wordtopo-wbe.txt  |  wh_a <- align/baseline-whisper-large-v3-crossattn-auto-buckeye-segA-5h-asotTrue-bs-5-en1.0-sil1.0-wordtopo-wbe.txt  |  owls_a <- align/baseline-owls-1B-180K-crossattn-auto-buckeye-segA-5h-asotTrue-bs-5-en1.0-sil1.0-wordtopo-wbe.txt  |  vx_a <- align/baseline-voxtral-selfattn-buckeye-segA-5h-asotTrue-bs-5-en1.0-sil1.0-wordtopo-wbe.txt
     & 1.0 & $116$ & $56$ & $171$ & $77$ & $52$ & $53$ & $53$ \\
    \hline\hline
    % (cont.): mms-fa <- align/wav2vec2ctc-fproj_out-prefixfwd-abl-buckeye-segA-5h-L2_grad-pertoken-asotTrue-bs-5-en0.0-zsk1.0-wordtopo-wbe.txt  |  whisper <- align/whisper-large-v3-logmel-charlev-spc-abl-buckeye-segA-5h-L2_grad-pertoken-asotTrue-bs-5-en0.0-zsk1.0-wordtopo-wbe.txt  |  owls <- align/owls-1B-180K-charlev-logmel-buckeye-segA-5h-L2_grad-pertoken-asotTrue-bs-5-en0.0-zsk1.0-wordtopo-wbe.txt  |  voxtral <- align/voxtral-charlevlogmel-buckeye-segA-5h-L2_grad-pertoken-asotTrue-bs-5-en0.0-zsk1.0-wordtopo-wbe.txt  |  wh_a <- align/baseline-whisper-large-v3-crossattn-auto-buckeye-segA-5h-asotTrue-bs-5-en0.0-zsk1.0-wordtopo-wbe.txt  |  owls_a <- align/baseline-owls-1B-180K-crossattn-auto-buckeye-segA-5h-asotTrue-bs-5-en0.0-zsk1.0-wordtopo-wbe.txt  |  vx_a <- align/baseline-voxtral-selfattn-buckeye-segA-5h-asotTrue-bs-5-en0.0-zsk1.0-wordtopo-wbe.txt
    \multirow{5}{*}{\makecell[l]{z-score \\ $\kappa = 1$}} & 0.0 & $142$ & $70$ & $263$ & $94$ & $65$ & $66$ & $64$ \\
    \cline{2-9}
    % (cont.): mms-fa <- align/wav2vec2ctc-fproj_out-prefixfwd-abl-buckeye-segA-5h-L2_grad-pertoken-asotTrue-bs-5-en0.25-zsk1.0-wordtopo-wbe.txt  |  whisper <- align/whisper-large-v3-logmel-charlev-spc-abl-buckeye-segA-5h-L2_grad-pertoken-asotTrue-bs-5-en0.25-zsk1.0-wordtopo-wbe.txt  |  owls <- align/owls-1B-180K-charlev-logmel-buckeye-segA-5h-L2_grad-pertoken-asotTrue-bs-5-en0.25-zsk1.0-wordtopo-wbe.txt  |  voxtral <- align/voxtral-charlevlogmel-buckeye-segA-5h-L2_grad-pertoken-asotTrue-bs-5-en0.25-zsk1.0-wordtopo-wbe.txt  |  wh_a <- align/baseline-whisper-large-v3-crossattn-auto-buckeye-segA-5h-asotTrue-bs-5-en0.25-zsk1.0-wordtopo-wbe.txt  |  owls_a <- align/baseline-owls-1B-180K-crossattn-auto-buckeye-segA-5h-asotTrue-bs-5-en0.25-zsk1.0-wordtopo-wbe.txt  |  vx_a <- align/baseline-voxtral-selfattn-buckeye-segA-5h-asotTrue-bs-5-en0.25-zsk1.0-wordtopo-wbe.txt
     & 0.25 & $137$ & $57$ & $210$ & $81$ & $63$ & $64$ & $59$ \\
    \cline{2-9}
    % (cont.): mms-fa <- align/wav2vec2ctc-fproj_out-prefixfwd-abl-buckeye-segA-5h-L2_grad-pertoken-asotTrue-bs-5-en0.5-zsk1.0-wordtopo-wbe.txt  |  whisper <- align/whisper-large-v3-logmel-charlev-spc-abl-buckeye-segA-5h-L2_grad-pertoken-asotTrue-bs-5-en0.5-zsk1.0-wordtopo-wbe.txt  |  owls <- align/owls-1B-180K-charlev-logmel-buckeye-segA-5h-L2_grad-pertoken-asotTrue-bs-5-en0.5-zsk1.0-wordtopo-wbe.txt  |  voxtral <- align/voxtral-charlevlogmel-buckeye-segA-5h-L2_grad-pertoken-asotTrue-bs-5-en0.5-zsk1.0-wordtopo-wbe.txt  |  wh_a <- align/baseline-whisper-large-v3-crossattn-auto-buckeye-segA-5h-asotTrue-bs-5-en0.5-zsk1.0-wordtopo-wbe.txt  |  owls_a <- align/baseline-owls-1B-180K-crossattn-auto-buckeye-segA-5h-asotTrue-bs-5-en0.5-zsk1.0-wordtopo-wbe.txt  |  vx_a <- align/baseline-voxtral-selfattn-buckeye-segA-5h-asotTrue-bs-5-en0.5-zsk1.0-wordtopo-wbe.txt
     & 0.5 & $127$ & $53$ & $186$ & $75$ & $61$ & $59$ & $56$ \\
    \cline{2-9}
    % (cont.): mms-fa <- align/wav2vec2ctc-fproj_out-prefixfwd-abl-buckeye-segA-5h-L2_grad-pertoken-asotTrue-bs-5-en0.75-zsk1.0-wordtopo-wbe.txt  |  whisper <- align/whisper-large-v3-logmel-charlev-spc-abl-buckeye-segA-5h-L2_grad-pertoken-asotTrue-bs-5-en0.75-zsk1.0-wordtopo-wbe.txt  |  owls <- align/owls-1B-180K-charlev-logmel-buckeye-segA-5h-L2_grad-pertoken-asotTrue-bs-5-en0.75-zsk1.0-wordtopo-wbe.txt  |  voxtral <- align/voxtral-charlevlogmel-buckeye-segA-5h-L2_grad-pertoken-asotTrue-bs-5-en0.75-zsk1.0-wordtopo-wbe.txt  |  wh_a <- align/baseline-whisper-large-v3-crossattn-auto-buckeye-segA-5h-asotTrue-bs-5-en0.75-zsk1.0-wordtopo-wbe.txt  |  owls_a <- align/baseline-owls-1B-180K-crossattn-auto-buckeye-segA-5h-asotTrue-bs-5-en0.75-zsk1.0-wordtopo-wbe.txt  |  vx_a <- align/baseline-voxtral-selfattn-buckeye-segA-5h-asotTrue-bs-5-en0.75-zsk1.0-wordtopo-wbe.txt
     & 0.75 & $120$ & $54$ & $175$ & $76$ & $57$ & $55$ & $54$ \\
    \cline{2-9}
    % (cont.): mms-fa <- align/wav2vec2ctc-fproj_out-prefixfwd-abl-buckeye-segA-5h-L2_grad-pertoken-asotTrue-bs-5-en1.0-zsk1.0-wordtopo-wbe.txt  |  whisper <- align/whisper-large-v3-logmel-charlev-spc-abl-buckeye-segA-5h-L2_grad-pertoken-asotTrue-bs-5-en1.0-zsk1.0-wordtopo-wbe.txt  |  owls <- align/owls-1B-180K-charlev-logmel-buckeye-segA-5h-L2_grad-pertoken-asotTrue-bs-5-en1.0-zsk1.0-wordtopo-wbe.txt  |  voxtral <- align/voxtral-charlevlogmel-buckeye-segA-5h-L2_grad-pertoken-asotTrue-bs-5-en1.0-zsk1.0-wordtopo-wbe.txt  |  wh_a <- align/baseline-whisper-large-v3-crossattn-auto-buckeye-segA-5h-asotTrue-bs-5-en1.0-zsk1.0-wordtopo-wbe.txt  |  owls_a <- align/baseline-owls-1B-180K-crossattn-auto-buckeye-segA-5h-asotTrue-bs-5-en1.0-zsk1.0-wordtopo-wbe.txt  |  vx_a <- align/baseline-voxtral-selfattn-buckeye-segA-5h-asotTrue-bs-5-en1.0-zsk1.0-wordtopo-wbe.txt
     & 1.0 & $117$ & $57$ & $173$ & $79$ & $52$ & $52$ & $54$ \\
    \hline
  \end{tabular}
  \end{adjustbox}
\end{table}

% =====================================================================
% AUTO-GENERATED by scripts/render_tables.py -- DO NOT EDIT THIS FILE.
% Regenerate from (edit these, then re-run render_tables.py / sync_tables_preview.sh):
%   tables-data/alignopts-dtw.data.json    -- the numbers (from the Sisyphus graph)
%   tables-spec/alignopts-dtw.spec.json     -- columns, units, layout
%   tables-spec/alignopts-dtw.caption.tex   -- caption text
% =====================================================================
% data source (Sisyphus recipe): i6_experiments/users/zeyer/experiments/exp2025_07_07_in_grads/jobs/grad_align_tables.py :: _alignopts_dtw_table
% per-row comments below map columns -> the registered output name the number comes from
\begin{table}[tb]
  \centering
  \caption{%
\textbf{Whisper's cross-attention DTW vs.~our aligner}
(Whisper-base cross-attention or grad.~scores, Buckeye).
Top row reproduces Whisper's \texttt{find\_alignment}.
Columns:
alignment-head set
(Whisper's curated heads, our gold-tuned top-$k$, or single best),
token z-norm, median filter,
pre-log of the score matrix (\emph{log}),
log-softmax over time (\emph{log sm}; DP sums these log-scores),
mono (\checkmark${}={}$our monotonic DP forbidding vertical step; $\times{}={}$Whisper DTW),
silence topology (\emph{word}${}={}$blank only between words; \emph{none}${}={}$no blank),
and energy weighting (\emph{en}).%
}

  \label{tab:alignopts-dtw}
  \begin{adjustbox}{max width=\linewidth}
  \begin{tabular}{|l|c|c|c|c|c|c|c|c|r|}
    \hline
    \textbf{Configuration} & \textbf{heads} & \textbf{\makecell{z-\\norm}} & \textbf{\makecell{med.\\filt.}} & \textbf{log} & \textbf{\makecell{log\\sm}} & \textbf{mono} & \textbf{sil.} & \textbf{en.} & \textbf{\makecell{WBE \\ {\normalfont {[ms]}\,$\downarrow$}}} \\ \hline\hline
    \multicolumn{10}{|l|}{\textbf{Cross-attn}} \\
    \hline
    % (cont.): wbe <- dtw-abl/whisper-base-buckeye-segA-5h-faithful-wbe.txt
    Whisper (faithful) & \multirow{9}{*}{\makecell[l]{Whi-\\sper}} & \multirow{5}{*}{\checkmark} & \multirow{4}{*}{\checkmark} & \multirow{7}{*}{$\times$} & \multirow{6}{*}{$\times$} & \multirow{2}{*}{$\times$} & \multirow{3}{*}{none} & $\times$ & $85$ \\
    \cline{1-1}\cline{9-10}
    % (cont.): wbe <- dtw-abl/whisper-base-buckeye-segA-5h-faithful_energy-wbe.txt
    \ $+$ energy &  &  &  &  &  &  &  & \checkmark & $89$ \\
    \cline{1-1}\cline{7-7}\cline{9-10}
    % (cont.): wbe <- dtw-abl/whisper-base-buckeye-segA-5h-faithful_mono-wbe.txt
    \ $+$ mono DP &  &  &  &  &  & \multirow{2}{*}{\checkmark} &  & \multirow{9}{*}{$\times$} & $84$ \\
    \cline{1-1}\cline{8-8}\cline{10-10}
    % (cont.): wbe <- dtw-abl/whisper-base-buckeye-segA-5h-faithful_silence-wbe.txt
    \ \quad $+$ silence &  &  &  &  &  &  & word &  & $84$ \\
    \cline{1-1}\cline{4-4}\cline{7-8}\cline{10-10}
    % (cont.): wbe <- dtw-abl/whisper-base-buckeye-segA-5h-nomedfilt-wbe.txt
    \ $+$ no median-filter &  &  & $\times$ &  &  & \multirow{3}{*}{$\times$} & \multirow{4}{*}{none} &  & $85$ \\
    \cline{1-1}\cline{3-4}\cline{10-10}
    % (cont.): wbe <- dtw-abl/whisper-base-buckeye-segA-5h-noznorm-wbe.txt
    \ $+$ no z-norm &  & \multirow{4}{*}{$\times$} & \multirow{6}{*}{\checkmark} &  &  &  &  &  & $83$ \\
    \cline{1-1}\cline{6-6}\cline{10-10}
    % (cont.): wbe <- dtw-abl/whisper-base-buckeye-segA-5h-noznorm_log-wbe.txt
    \ \quad $+$ log softmax &  &  &  &  & \multirow{3}{*}{\checkmark} &  &  &  & $81$ \\
    \cline{1-1}\cline{5-5}\cline{7-7}\cline{10-10}
    % (cont.): wbe <- dtw-abl/whisper-base-buckeye-segA-5h-noznorm_log_mono-wbe.txt
    \ \qquad $+$ mono DP &  &  &  & \multirow{2}{*}{\checkmark} &  & \multirow{2}{*}{\checkmark} &  &  & $83$ \\
    \cline{1-1}\cline{8-8}\cline{10-10}
    % (cont.): wbe <- dtw-abl/whisper-base-buckeye-segA-5h-noznorm_log_mono_sil-wbe.txt
    \ \qquad\quad $+$ silence &  &  &  &  &  &  & word &  & $76$ \\
    \cline{1-3}\cline{5-8}\cline{10-10}
    % (cont.): wbe <- dtw-abl/whisper-base-buckeye-segA-5h-ourheads-wbe.txt
    \ $+$ our heads & ours & \multirow{2}{*}{\checkmark} &  & \multirow{2}{*}{$\times$} & \multirow{2}{*}{$\times$} & \multirow{2}{*}{$\times$} & \multirow{2}{*}{none} &  & $79$ \\
    \cline{1-2}\cline{10-10}
    % (cont.): wbe <- dtw-abl/whisper-base-buckeye-segA-5h-best1head-wbe.txt
    \ $+$ single best head & 1-best &  &  &  &  &  &  &  & $95$ \\
    \hline
    % (cont.): wbe <- dtw-abl/whisper-base-buckeye-segA-5h-ours_full-wbe.txt
    ours (full) & \multirow{4}{*}{ours} & \multirow{4}{*}{$\times$} & \multirow{4}{*}{$\times$} & \checkmark & \multirow{4}{*}{\checkmark} & \checkmark & word & \multirow{2}{*}{\checkmark} & $71$ \\
    \cline{1-1}\cline{5-5}\cline{7-8}\cline{10-10}
    % (cont.): wbe <- dtw-abl/whisper-base-buckeye-segA-5h-ours_dtw-wbe.txt
    \ $+$ DTW &  &  &  & \multirow{2}{*}{$\times$} &  & \multirow{2}{*}{$\times$} & \multirow{3}{*}{none} &  & $73$ \\
    \cline{1-1}\cline{9-10}
    % (cont.): wbe <- dtw-abl/whisper-base-buckeye-segA-5h-ours_dtw_noen-wbe.txt
    \ \quad $+$ no energy &  &  &  &  &  &  &  & $\times$ & $71$ \\
    \cline{1-1}\cline{5-5}\cline{7-7}\cline{9-10}
    % (cont.): wbe <- dtw-abl/whisper-base-buckeye-segA-5h-ours_none-wbe.txt
    \ $+$ no silence &  &  &  & \checkmark &  & \checkmark &  & \checkmark & $117$ \\
    \hline\hline
    \multicolumn{10}{|l|}{\textbf{Grad}} \\
    \hline
    % (cont.): wbe <- align/whisper-base-logmel-buckeye-segA-5h-L2_grad-pertoken-charlev-spc-asotTrue-bs-5-en0.5-sil2.0-wordtopo-wbe.txt
    ours (full) & \multirow{4}{*}{--} & \multirow{4}{*}{--} & \multirow{4}{*}{--} & \multirow{4}{*}{\checkmark} & \multirow{4}{*}{\checkmark} & \checkmark & \multirow{2}{*}{word} & \multirow{3}{*}{\checkmark} & $75$ \\
    \cline{1-1}\cline{7-7}\cline{10-10}
    % (cont.): wbe <- align/whisper-base-logmel-buckeye-segA-5h-L2_grad-pertoken-charlev-spc-asotTrue-bs-5-en0.5-sil2.0-dtwword-wbe.txt
    \ $+$ DTW, word-topo &  &  &  &  &  & \multirow{3}{*}{$\times$} &  &  & $76$ \\
    \cline{1-1}\cline{8-8}\cline{10-10}
    % (cont.): wbe <- align/whisper-base-logmel-buckeye-segA-5h-L2_grad-pertoken-charlev-spc-asotTrue-bs-5-en0.5-truedtw-wbe.txt
    \ \quad $+$ no word-topo &  &  &  &  &  &  & \multirow{2}{*}{none} &  & $138$ \\
    \cline{1-1}\cline{9-10}
    % (cont.): wbe <- align/whisper-base-logmel-buckeye-segA-5h-L2_grad-pertoken-charlev-spc-asotTrue-bs-5-en0.0-truedtw-wbe.txt
    \ \qquad $+$ no energy &  &  &  &  &  &  &  & $\times$ & $152$ \\
    \hline
  \end{tabular}
  \end{adjustbox}
\end{table}

% =====================================================================
% AUTO-GENERATED by scripts/render_tables.py -- DO NOT EDIT THIS FILE.
% Regenerate from (edit these, then re-run render_tables.py / sync_tables_preview.sh):
%   tables-data/wav2vec-resolution.data.json    -- the numbers (from the Sisyphus graph)
%   tables-spec/wav2vec-resolution.spec.json     -- columns, units, layout
%   tables-spec/wav2vec-resolution.caption.tex   -- caption text
% =====================================================================
% data source (Sisyphus recipe): i6_experiments/users/zeyer/experiments/exp2025_07_07_in_grads/jobs/grad_align_tables.py :: _wav2vec_resolution_table
% per-row comments below map columns -> the registered output name the number comes from
\begin{table}[tb]
  \centering
  \caption{%
\textbf{MMS-FA (wav2vec 2.0) CTC gradient alignment across the internal level the gradient is taken at}, Buckeye.
The levels run from the convolutional feature encoder through the feature projection to the raw waveform at several pooling rates.
The feat-proj linear row equals the MMS-FA gradient number reported in \cref{tab:per-model-methods}.%
}

  \label{tab:wav2vec-resolution}
  \begin{adjustbox}{max width=\linewidth}
  \begin{tabular}{|l|r|r|r|r|}
    \hline
    \textbf{Gradient w.r.t.} & \textbf{ms/frame} & \textbf{\makecell{Grid \\ {\normalfont {[Hz]}}}} & \textbf{\makecell{WBE \\ {\normalfont {[ms]}\,$\downarrow$}}} & \textbf{\makecell{$\le$50ms \\ {\normalfont {[\%]}\,$\uparrow$}}} \\ \hline\hline
    % (cont.): wbe,a50 <- align/wav2vec2ctc-conv0-buckeye-segA-5h-L2_grad-pertoken-asotTrue-bs-5-en0.5-sil2.0-wordtopo-wbe.txt
    Conv 0 & 0.31 & 3200 & $137$ & $48.7$ \\
    \hline
    % (cont.): wbe,a50 <- align/wav2vec2ctc-conv1-buckeye-segA-5h-L2_grad-pertoken-asotTrue-bs-5-en0.5-sil2.0-wordtopo-wbe.txt
    Conv 1 & 0.62 & 1600 & $133$ & $49.5$ \\
    \hline
    % (cont.): wbe,a50 <- align/wav2vec2ctc-conv2-buckeye-segA-5h-L2_grad-pertoken-asotTrue-bs-5-en0.5-sil2.0-wordtopo-wbe.txt
    Conv 2 & 1.25 & 800 & $133$ & $49.5$ \\
    \hline
    % (cont.): wbe,a50 <- align/wav2vec2ctc-conv3-buckeye-segA-5h-L2_grad-pertoken-asotTrue-bs-5-en0.5-sil2.0-wordtopo-wbe.txt
    Conv 3 & 2.5 & 400 & $132$ & $50.0$ \\
    \hline
    % (cont.): wbe,a50 <- align/wav2vec2ctc-conv4-buckeye-segA-5h-L2_grad-pertoken-asotTrue-bs-5-en0.5-sil2.0-wordtopo-wbe.txt
    Conv 4 & 5 & 200 & $130$ & $50.8$ \\
    \hline
    % (cont.): wbe,a50 <- align/wav2vec2ctc-conv5-buckeye-segA-5h-L2_grad-pertoken-asotTrue-bs-5-en0.5-sil2.0-wordtopo-wbe.txt
    Conv 5 & 10 & 100 & $128$ & $52.3$ \\
    \hline
    % (cont.): wbe,a50 <- align/wav2vec2ctc-fproj_ln-buckeye-segA-5h-L2_grad-pertoken-asotTrue-bs-5-en0.5-sil2.0-wordtopo-wbe.txt
    Feat-proj LayerNorm & 20 & 50 & $122$ & $55.3$ \\
    \hline
    % (cont.): wbe,a50 <- align/wav2vec2ctc-fproj_out-buckeye-segA-5h-L2_grad-pertoken-asotTrue-bs-5-en0.5-sil2.0-wordtopo-wbe.txt
    Feat-proj Linear & 20 & 50 & $121$ & $58.2$ \\
    \hline
    % (cont.): wbe,a50 <- align/wav2vec2ctc-rawwav-buckeye-segA-5h-L2_grad-pertoken-asotTrue-bs-5-en0.5-sil2.0-wordtopo-wbe.txt
    Raw waveform, pool 320 & 20 & 50 & $135$ & $52.1$ \\
    \hline
    % (cont.): wbe,a50 <- align/wav2vec2ctc-rawwav-pool80-buckeye-segA-5h-L2_grad-pertoken-asotTrue-bs-5-en0.5-sil2.0-wordtopo-wbe.txt
    Raw waveform, pool 80 & 5 & 200 & $140$ & $48.4$ \\
    \hline
    % (cont.): wbe,a50 <- align/wav2vec2ctc-rawwav-pool16-buckeye-segA-5h-L2_grad-pertoken-asotTrue-bs-5-en0.5-sil2.0-wordtopo-wbe.txt
    Raw waveform, pool 16 & 1 & 1000 & $142$ & $47.5$ \\
    \hline
    % (cont.): wbe,a50 <- align/wav2vec2ctc-rawwav-pool1-buckeye-segA-5h-L2_grad-pertoken-asotTrue-bs-5-en0.5-sil2.0-wordtopo-wbe.txt
    Raw waveform, pool 1 & 0.0625 & 16000 & $134$ & $47.7$ \\
    \hline
  \end{tabular}
  \end{adjustbox}
\end{table}

% =====================================================================
% AUTO-GENERATED by scripts/render_tables.py -- DO NOT EDIT THIS FILE.
% Regenerate from (edit these, then re-run render_tables.py / sync_tables_preview.sh):
%   tables-data/encoder-depth.data.json    -- the numbers (from the Sisyphus graph)
%   tables-spec/encoder-depth.spec.json     -- columns, units, layout
%   tables-spec/encoder-depth.caption.tex   -- caption text
% =====================================================================
% data source (Sisyphus recipe): i6_experiments/users/zeyer/experiments/exp2025_07_07_in_grads/jobs/grad_align_tables.py :: _encoder_depth_table
% per-row comments below map columns -> the registered output name the number comes from
\begin{table}[tb]
  \centering
  \caption{%
\textbf{Encoder depth the gradient is taken at} for gradient alignment.
Buckeye.
Rows run from log-mel input through encoder to its output.
Whisper-large-v3 uses char targets (32 layers; 1/4 etc.~= L8/16/24)
and FastConformer-CTC subword targets (17 layers; L4/9/13).
Off.~is mean signed boundary offset (positive = late).%
}

  \label{tab:encoder-depth}
  \begin{adjustbox}{max width=\linewidth}
  \begin{tabular}{|l|r|r|r|r|r|r|r|r|}
    \hline
    \multirow{2}{*}{\textbf{\makecell[l]{Gradient\\w.r.t.}}} & \multicolumn{4}{c|}{\textbf{Whisper-large-v3 (char)}} & \multicolumn{4}{c|}{\textbf{FastConformer-CTC}} \\
    \cline{2-5} \cline{6-9}
     & \textbf{\makecell{Grid \\ {\normalfont {[Hz]}}}} & \textbf{\makecell{WBE \\ {\normalfont {[ms]}\,$\downarrow$}}} & \textbf{\makecell{$\le$50ms \\ {\normalfont {[\%]}\,$\uparrow$}}} & \textbf{\makecell{Off. \\ {\normalfont {[ms]}}}} & \textbf{\makecell{Grid \\ {\normalfont {[Hz]}}}} & \textbf{\makecell{WBE \\ {\normalfont {[ms]}\,$\downarrow$}}} & \textbf{\makecell{$\le$50ms \\ {\normalfont {[\%]}\,$\uparrow$}}} & \textbf{\makecell{Off. \\ {\normalfont {[ms]}}}} \\ \hline\hline
    % (cont.): w_wbe,w_a50,w_off <- align/whisper-large-v3-charlev-spc-logmel-encdepth-buckeye-segA-5h-L2_grad-pertoken-asotTrue-bs-5-en0.5-sil2.0-wordtopo-wbe.txt  |  fc_wbe,fc_a50,fc_off <- align/fastconformer-stream-ctc-logmel-encdepth-buckeye-segA-5h-L2_grad-pertoken-asotTrue-bs-5-en0.5-sil2.0-wordtopo-wbe.txt
    Log-mel in & $100$ & $53$ & $66.2$ & $-10$ & $100$ & $158$ & $30.1$ & $-80$ \\
    \hline
    % (cont.): w_wbe,w_a50,w_off <- align/whisper-large-v3-charlev-spc-encin-encdepth-buckeye-segA-5h-L2_grad-pertoken-asotTrue-bs-5-en0.5-sil2.0-wordtopo-wbe.txt  |  fc_wbe,fc_a50,fc_off <- align/fastconformer-stream-ctc-encin-encdepth-buckeye-segA-5h-L2_grad-pertoken-asotTrue-bs-5-en0.5-sil2.0-wordtopo-wbe.txt
    Encoder in & \multirow{5}{*}{$50$} & $44$ & $74.6$ & $-9$ & \multirow{5}{*}{$12.5$} & $152$ & $30.0$ & $-63$ \\
    \cline{1-1}\cline{3-5}\cline{7-9}
    % (cont.): w_wbe,w_a50,w_off <- align/whisper-large-v3-charlev-spc-encL8-encdepth-buckeye-segA-5h-L2_grad-pertoken-asotTrue-bs-5-en0.5-sil2.0-wordtopo-wbe.txt  |  fc_wbe,fc_a50,fc_off <- align/fastconformer-stream-ctc-encL4-encdepth-buckeye-segA-5h-L2_grad-pertoken-asotTrue-bs-5-en0.5-sil2.0-wordtopo-wbe.txt
    Encoder 1/4 &  & $42$ & $76.3$ & $-9$ &  & $145$ & $29.4$ & $-33$ \\
    \cline{1-1}\cline{3-5}\cline{7-9}
    % (cont.): w_wbe,w_a50,w_off <- align/whisper-large-v3-charlev-spc-encL16-encdepth-buckeye-segA-5h-L2_grad-pertoken-asotTrue-bs-5-en0.5-sil2.0-wordtopo-wbe.txt  |  fc_wbe,fc_a50,fc_off <- align/fastconformer-stream-ctc-encL9-encdepth-buckeye-segA-5h-L2_grad-pertoken-asotTrue-bs-5-en0.5-sil2.0-wordtopo-wbe.txt
    Encoder 1/2 &  & $41$ & $78.2$ & $-9$ &  & $146$ & $23.9$ & $+73$ \\
    \cline{1-1}\cline{3-5}\cline{7-9}
    % (cont.): w_wbe,w_a50,w_off <- align/whisper-large-v3-charlev-spc-encL24-encdepth-buckeye-segA-5h-L2_grad-pertoken-asotTrue-bs-5-en0.5-sil2.0-wordtopo-wbe.txt  |  fc_wbe,fc_a50,fc_off <- align/fastconformer-stream-ctc-encL13-encdepth-buckeye-segA-5h-L2_grad-pertoken-asotTrue-bs-5-en0.5-sil2.0-wordtopo-wbe.txt
    Encoder 3/4 &  & $39$ & $80.0$ & $-9$ &  & $229$ & $12.5$ & $+196$ \\
    \cline{1-1}\cline{3-5}\cline{7-9}
    % (cont.): w_wbe,w_a50,w_off <- align/whisper-large-v3-charlev-spc-encout-encdepth-buckeye-segA-5h-L2_grad-pertoken-asotTrue-bs-5-en0.5-sil2.0-wordtopo-wbe.txt  |  fc_wbe,fc_a50,fc_off <- align/fastconformer-stream-ctc-encout-encdepth-buckeye-segA-5h-L2_grad-pertoken-asotTrue-bs-5-en0.5-sil2.0-wordtopo-wbe.txt
    Encoder out &  & $42$ & $79.0$ & $-13$ &  & $305$ & $4.9$ & $+274$ \\
    \hline
  \end{tabular}
  \end{adjustbox}
\end{table}

% =====================================================================
% AUTO-GENERATED by scripts/render_tables.py -- DO NOT EDIT THIS FILE.
% Regenerate from (edit these, then re-run render_tables.py / sync_tables_preview.sh):
%   tables-data/owsm-per-layer.data.json    -- the numbers (from the Sisyphus graph)
%   tables-spec/owsm-per-layer.spec.json     -- columns, units, layout
%   tables-spec/owsm-per-layer.caption.tex   -- caption text
% =====================================================================
% data source (Sisyphus recipe): i6_experiments/users/zeyer/experiments/exp2025_07_07_in_grads/jobs/grad_align_tables.py :: _owsm_layer_table
% per-row comments below map columns -> the registered output name the number comes from
\begin{table}[tb]
  \centering
  \caption{%
\textbf{OWSM-CTC alignment from each inter-CTC block} (6/12/15/21) and the final block (27).
%WBE on TIMIT-test and Buckeye.
Gradient alignment vs.~posterior alignment.%
}

  \label{tab:owsm-per-layer}
  \begin{adjustbox}{max width=\linewidth}
  \begin{tabular}{|l|r|r|r|r|}
    \hline
    \multirow{3}{*}{\textbf{Emit block}} & \multicolumn{4}{c|}{\textbf{WBE\,{\normalfont {[ms]}\,$\downarrow$}}} \\
    \cline{2-5}
     & \multicolumn{2}{c|}{\textbf{TIMIT}} & \multicolumn{2}{c|}{\textbf{Buckeye}} \\
    \cline{2-3} \cline{4-5}
     & \textbf{Grad.} & \textbf{Post.} & \textbf{Grad.} & \textbf{Post.} \\ \hline\hline
    % 6: t_grad <- align/owsm-ctc-v4-1b-lyr6-timit-test-L2_grad-pertoken-asotTrue-bs-5-en0.5-sil2.0-wordtopo-wbe.txt  |  t_fa <- baseline-owsm-ctc-v4-1b-lyr6-timit-test-wbe.txt  |  s_grad <- align/owsm-ctc-v4-1b-lyr6-buckeye-segA-5h-L2_grad-pertoken-asotTrue-bs-5-en0.5-sil2.0-wordtopo-wbe.txt  |  s_fa <- baseline-owsm-ctc-v4-1b-lyr6-buckeye-segA-5h-wbe.txt
    6 & $142$ & $120$ & $145$ & $110$ \\
    \hline
    % 12: t_grad <- align/owsm-ctc-v4-1b-lyr12-timit-test-L2_grad-pertoken-asotTrue-bs-5-en0.5-sil2.0-wordtopo-wbe.txt  |  t_fa <- baseline-owsm-ctc-v4-1b-lyr12-timit-test-wbe.txt  |  s_grad <- align/owsm-ctc-v4-1b-lyr12-buckeye-segA-5h-L2_grad-pertoken-asotTrue-bs-5-en0.5-sil2.0-wordtopo-wbe.txt  |  s_fa <- baseline-owsm-ctc-v4-1b-lyr12-buckeye-segA-5h-wbe.txt
    12 & $150$ & $124$ & $167$ & $118$ \\
    \hline
    % 15: t_grad <- align/owsm-ctc-v4-1b-lyr15-timit-test-L2_grad-pertoken-asotTrue-bs-5-en0.5-sil2.0-wordtopo-wbe.txt  |  t_fa <- baseline-owsm-ctc-v4-1b-lyr15-timit-test-wbe.txt  |  s_grad <- align/owsm-ctc-v4-1b-lyr15-buckeye-segA-5h-L2_grad-pertoken-asotTrue-bs-5-en0.5-sil2.0-wordtopo-wbe.txt  |  s_fa <- baseline-owsm-ctc-v4-1b-lyr15-buckeye-segA-5h-wbe.txt
    15 & $160$ & $125$ & $183$ & $119$ \\
    \hline
    % 21: t_grad <- align/owsm-ctc-v4-1b-lyr21-timit-test-L2_grad-pertoken-asotTrue-bs-5-en0.5-sil2.0-wordtopo-wbe.txt  |  t_fa <- baseline-owsm-ctc-v4-1b-lyr21-timit-test-wbe.txt  |  s_grad <- align/owsm-ctc-v4-1b-lyr21-buckeye-segA-5h-L2_grad-pertoken-asotTrue-bs-5-en0.5-sil2.0-wordtopo-wbe.txt  |  s_fa <- baseline-owsm-ctc-v4-1b-lyr21-buckeye-segA-5h-wbe.txt
    21 & $168$ & $125$ & $194$ & $120$ \\
    \hline
    % 27 (final): t_grad <- align/owsm-ctc-v4-1b-prefixfwd-timit-test-L2_grad-pertoken-asotTrue-bs-5-en0.5-sil2.0-wordtopo-wbe.txt  |  t_fa <- baseline-owsm-ctc-v4-1b-timit-test-wbe.txt  |  s_grad <- align/owsm-ctc-v4-1b-prefixfwd-buckeye-segA-5h-L2_grad-pertoken-asotTrue-bs-5-en0.5-sil2.0-wordtopo-wbe.txt  |  s_fa <- baseline-owsm-ctc-v4-1b-buckeye-segA-5h-wbe.txt
    27 (final) & $172$ & $125$ & $198$ & $121$ \\
    \hline
  \end{tabular}
  \end{adjustbox}
\end{table}

% =====================================================================
% AUTO-GENERATED by scripts/render_tables.py -- DO NOT EDIT THIS FILE.
% Regenerate from (edit these, then re-run render_tables.py / sync_tables_preview.sh):
%   tables-data/cost.data.json    -- the numbers (from the Sisyphus graph)
%   tables-spec/cost.spec.json     -- columns, units, layout
%   tables-spec/cost.caption.tex   -- caption text
% =====================================================================
% data source (Sisyphus recipe): i6_experiments/users/zeyer/experiments/exp2025_07_07_in_grads/jobs/grad_align_tables.py :: _cost_table
% per-row comments below map columns -> the registered output name the number comes from
\begin{table}[tb]
  \centering
  \caption{%
\textbf{Cost of gradient alignment}, single GPU, subset of TIMIT,
as real-time factor (RTF) $\times 10^3$ (ms compute per second of audio).
Stages: Forward (encoder $\rightarrow$ logits, shared with native methods),
prefix-sum (CTC/transducer only),
and backward (encoder VJP, batched).
Grad.~align extra cost is prefix-sum $+$ backward;
%AED and the speech LLM have no lattice (prefix-sum n/a).
Shared DP align is excluded.%
}

  \label{tab:cost}
  \begin{adjustbox}{max width=\linewidth}
  \begin{tabular}{|l|l|r|r|r|r|}
    \hline
    \multicolumn{2}{|c|}{\textbf{Model}} & \multicolumn{4}{c|}{\textbf{Compute time\,{\normalfont {[RTF$\times 10^3$]}}}} \\
    \cline{1-2} \cline{3-6}
    \multirow{2}{*}{\textbf{Type}} & \multirow{2}{*}{\textbf{Name}} & \multicolumn{2}{c|}{\textbf{Model}} & \multicolumn{2}{c|}{\textbf{Prefix-sum}} \\
    \cline{3-4} \cline{5-6}
     &  & \textbf{Fwd} & \textbf{Bwd} & \textbf{Fwd} & \textbf{Bwd} \\ \hline\hline
    % (cont.): forward,backward,prefix_fwd,prefix_bwd <- speedcmp-p212/cost-4way-metrics.txt
    CTC & MMS-FA & $6$ & $57$ & $3$ & $212$ \\
    \hline
    % (cont.): forward,backward,prefix_fwd,prefix_bwd <- speedcmp-p212/cost-4way-metrics.txt
    AED & Whisper-large-v3 & $24$ & $277$ & \multicolumn{2}{c|}{\multirow{2}{*}{n/a}} \\
    \cline{1-4}
    % (cont.): forward,backward,prefix_fwd,prefix_bwd <- speedcmp-p212/cost-4way-metrics.txt
    Speech LLM & Phi-4-MM & $26$ & $216$ & \multicolumn{2}{c|}{} \\
    \hline
  \end{tabular}
  \end{adjustbox}
\end{table}

\section{Experimental Setup}

We evaluate sixteen models from the four families.
For reference we run two dedicated aligners,
a GM-HMM via MFA~\cite{mcauliffe2017montreal},
and MMS-FA~\cite{pratap2024mms,hwang2023torchaudio},
a {wav2vec 2.0}~\cite{baevski2020wav2vec} CTC model built specifically for forced alignment.
For CTC we further use XLS-R on phonemes~\cite{phy22-phoneme,babu2022xlsr}, Parakeet CTC~\cite{rekesh2023fastconformer}, OWSM-CTC~\cite{peng2025owsmv4,peng2024owsmctc},
and a streaming FastConformer-CTC~\cite{Noroozi2024CacheAwareFastConformer}.
For the transducers we use Parakeet RNN-T~\cite{rekesh2023fastconformer} and TDT~\cite{xu2023tdt}, and the streaming FastConformer-RNN-T~\cite{Noroozi2024CacheAwareFastConformer} and Emformer~\cite{shi2021emformer,yang2022torchaudio}.
The AED models are Whisper base and large-v3~\cite{radford2023whisper}, CrisperWhisper~\cite{wagner2024crisperwhisper}, and OWLS-1B~\cite{chen2025owls},
and the speech LLMs are Voxtral~\cite{mistral2025voxtral}, Phi-4-multimodal (Phi-4-MM)~\cite{microsoft2025phi4mini}, and Canary-Qwen~\cite{puvvada2024canary}.
We align on TIMIT (test)~\cite{garofolo1993timit} and on Buckeye spontaneous speech~\cite{pitt2005buckeye}, both with gold word boundaries.
For Buckeye we use a $5$\,h subset stratified by speaker (all speakers represented proportionally),
split at every $\geq 1$\,s inter-word silence,
then split any piece still longer than $18$\,s at its largest internal gap%
\footnote{\cite{huang2024lesspeaky} drops long utterances, but we keep them by splitting.}.
%
%\paragraph{Buckeye segmentation variants}
%\input{tables/buckeye-abc.tex}
%Across the three segmentation variants (\cref{tab:buckeye-abc}),
%the within-model ranking of the gradient against the native aligner is stable,
%and every method degrades together on the harder no-reseg variant,
%so the comparison is not an artifact of the segmentation.
%
We report the word-boundary error (WBE),
i.e.~the mean per-word start and end absolute error, averaged over all words in the corpus%
\footnote{\cite{rousso24_interspeech} averages first per utterance and then over utterances.},
and the accuracy at a collar, i.e.~the fraction of all boundaries within $50$\,ms over the corpus.
We compare each model against its own alignment,
i.e.~the CTC or transducer forced alignment on the model's own emission (the ``posteriors''),
the AED cross-attention DTW~\cite{bain2023whisperx},
and, for the speech LLMs, the analogous text-to-audio self-attention,
decoded by the same dynamic program as the gradient scores.
We select the attention heads by per-head WBE on TIMIT development gold and average the top 8 heads.
In hypothesis mode, where the recognition differs from the reference,
we match the words by identity and report an identity-gated F1 at a $50$\,ms collar together with the matched WBE.

\section{Results}

\paragraph{Alignment quality across model families}

\Cref{tab:per-model-methods} compares the
gradient alignment versus each model's own native or attention aligner,
across CTC, AED, transducer and speech-LLM families.
Gradient alignment is competitive on every family,
beats the native alignment where it is weak (the streaming models, Canary-Qwen),
and at the starred best encoder depth it also beats the Whisper-large-v3 cross-attention.
The streaming posteriors' large error is largely a systematic emission-delay bias rather than scatter
(\cref{tab:boundary-offsets,fig:input-grad});
the same table shows that gradient alignment shifts both boundaries together (a positional lead)
rather than shrinking the word like the posteriors do.

\Cref{tab:hyp} aligns each model's own recognition
(identity-gated F1 at a $50$\,ms collar and matched WBE),
showing that gradient alignment remains usable when the hypothesis differs from the reference.

\paragraph{Tokenization and the attention baseline}

\Cref{tab:char-vs-subword-family} justifies the tokenization we use throughout:
the gradient alignment uses character targets for the AED and speech-LLM families
and subword targets for CTC and transducers,
while every attention or posterior baseline uses the model's native subword tokens.
Character targets improve the gradient markedly for the autoregressive families
(on Whisper-large-v3 and Voxtral the char gradient is well below the subword gradient),
but they do not help the model-native attention aligners,
and for the non-autoregressive CTC and transducer, which have no per-character acoustic emission,
char-level collapses: char targets are consistently worse than subword for both the gradient and the posteriors, in the worst cases by an order of magnitude.

\paragraph{Ablations}

The grad.~score per-token reduction (\cref{tab:gradscore})
moves WBE by only a few ms across the $p$-norms, while the plain sum is clearly worse,
and gradient versus gradient-$\times$-input~\cite{shrikumar2017learning,ancona2018towards} is within $\sim$1--3\,ms with neither winning consistently.
Multi-pass attribution (SmoothGrad~\cite{smilkov2017smoothgrad}, VarGrad~\cite{adebayo2018sanity}, Integrated~\cite{sundararajan2017axiomatic} and Expected Gradients~\cite{erion2021improving})
does not beat the single-pass gradient (numbers omitted here). %(\cref{tab:attribution}).
Most decoding options behave consistently (\cref{tab:alignopts-silence,tab:alignopts-energy,tab:alignopts-dtw}):
the gradient is active even in silence and needs the energy-aware ($\lambda\approx 2$) or z-score blank,
whereas the attention is already silence-aware and a constant blank suffices;
the z-score blank is the energy-free option that matches the energy one for gradients.
The audio-energy token-weighting exponent matters little around our default
and behaves consistently across the blank schemes (\cref{tab:alignopts-energy}).
For the cross-attention DTW, the dominant gain over the original Whisper heuristic
is the gold-tuned head selection (its curated set includes some near-useless heads);
the Whisper-DTW setting is exactly our decoder
with the log-compression and the blank state removed,
and energy weighting helps the gradient but not the cross-attention DTW.

\paragraph{Where the gradient is taken}

For MMS-FA, which is a wav2vec 2.0 model~\cite{baevski2020wav2vec} (\cref{tab:wav2vec-resolution}),
the gradient localizes best at the feature-projection output (20\,ms encoder grid);
finer levels (the convolutional feature encoder, or the raw waveform pooled down to sample resolution)
do not help, and the raw-waveform gradient is noisier.
\Cref{tab:encoder-depth} sweeps the depth for Whisper-large-v3 and the FastConformer-CTC,
from the log-mel input through the encoder to its output (the activation just before the decoder / CTC linear):
the sharpest gradient is at an intermediate depth
(three quarters of the Whisper encoder, the L24 of the starred row in \cref{tab:per-model-methods};
a quarter for the FastConformer-CTC),
not the input,
and in the streaming FastConformer-CTC the encoder's time shift accumulates with depth,
so its output gradient ends as delayed as its posteriors (\cref{fig:input-grad}).

\paragraph{Speech-LLM prompt influence}

Gradient alignment is robust to the prompt used across all the speech LLMs
and across the gradient-vs-self-attention signal
(numbers omitted).

\paragraph{Inter-CTC}

OWSM-CTC has intermediate CTC losses~\cite{peng2025owsmv4,peng2024owsmctc},
which we can use to take the gradient at each block of the encoder (\cref{tab:owsm-per-layer}).
Interestingly, the earlier blocks are better than the later ones,
both for the gradient and for the native CTC alignment.

\paragraph{Time stretch}

Under audio time-stretch (numbers omitted), % (\cref{tab:time-stretch}),
the same upsampling mechanism acts with opposite sign:
it hurts a fine-grid model (MMS-FA) but helps a coarse-grid one (Voxtral) at mild factors.

\paragraph{Compute cost}
The only extra cost over the native methods is the per-token backward (\cref{tab:cost});
the forward and the DP decode are shared,
and for the CTC the per-token prefix-score lattice backward dominates.

\section{Conclusion}
Gradient alignment produces a usable word alignment for every model and family we tried,
including the speech LLMs that have no built-in word aligner.
It is usually a little behind a strong native aligner,
but clearly better where that alignment is weak, as on the streaming models and Canary-Qwen,
and on Whisper-large-v3 it even beats its cross-attention at the best encoder depth.
Our decoder also improves on Whisper's cross-attention DTW itself (\cref{tab:alignopts-dtw}).
Its input-grid resolution lets finer character targets improve accuracy,
which the coarse attention grid cannot exploit.
The per-token backward makes it much more expensive than a single forward,
so we do not propose it as a practical aligner.
The result is rather that one training-free input saliency aligns every differentiable ASR model
across all four families, in some cases even better than the model's own alignment.
Beyond alignment, the gradient also serves as an analysis tool.

\FloatBarrier  % flush any pending tables before leaving the main content
\newpage

\if\blind0
\section*{Acknowledgment}

This work was partially supported by NeuroSys,
which as part of the initiative “Clusters4Future” is funded by the Federal Ministry of Research,
Technology and Space BMFTR (funding IDs 03ZU2106DA and 03ZU2106DD),
and by the project RESCALE within the program \textit{AI Lighthouse Projects for the Environment, Climate, Nature and Resources}
funded by the Federal Ministry for the Environment, Nature Conservation, Nuclear Safety and Consumer Protection (BMUV), funding ID: 67KI32006A.
The authors gratefully acknowledge the computing time provided to them
at the NHR Center NHR4CES at RWTH Aachen University
(project number p0023999).
% not really p0023565 here...
This is funded by the Federal Ministry of Education and Research, 
and the state governments participating on the basis 
of the resolutions of the GWK for 
national high performance computing at universities 
(\url{www.nhr-verein.de/unsere-partner}).

\fi

\section*{AI-Generated Content Disclosure}

% The use of content generated by artificial intelligence (AI) in an article (including but not limited to text, figures, images, and code) shall be disclosed in the acknowledgments section of any article submitted to an IEEE publication. The AI system used shall be identified, and specific sections of the article that use AI-generated content shall be identified and accompanied by a brief explanation regarding the level at which the AI system was used to generate the content. The use of AI systems for editing and grammar enhancement is common practice and, as such, is generally outside the intent of the above policy. In this case, disclosure as noted above is recommended.

% The AI-Generated Content Disclosure statement does not need to be included within the 6-page limit for regular and special session papers, or within the 3-page limit for demo and challenge papers. It may be placed after the main content of the paper and before the references.

% AI-generated content was used and will be disclosed in the acknowledgments in the camera-ready version (system + sections + brief level of use).

We used AI assistants (large language models) substantially in this work,
to help set up and run the experiments, to debug and write parts of the code,
and to draft and edit parts of this paper.
All results were verified by the authors,
who take full responsibility for the content.

% section References

\bibliographystyle{IEEEtran}
\bibliography{main}

@string{PROC-IEEE = "Proc.\ of the IEEE"}

@string{ICASSP = "Proc.\ IEEE ICASSP"}

@string{ICML = "Proc.\ ICML"}

@string{NEURIPS = "Proc.\ NeurIPS"}

@string{ASRU = "Proc.\ IEEE ASRU"}

@string{INTERSPEECH = "Proc.\ Interspeech"}

@string{NIPS = "Proc.\ NIPS"}

@string{SPEECHCOMMUNICATION = "Speech Communication"}

@string{ACL = "Proc.\ ACL"}

@string{EMNLP = "Proc.\ EMNLP"}

@string{JMLR = "Journal of Machine Learning Research"}

@string{ICLR = "Proc.\ ICLR"}

@string{TASLP = "IEEE/ACM Trans.\ Audio, Speech, and Language Processing"}

@inproceedings{graves2006ctc,
  title={Connectionist temporal classification: labelling unsegmented sequence data with recurrent neural networks},
  author={Graves, Alex and Fern{\'a}ndez, Santiago and Gomez, Faustino and Schmidhuber, J{\"u}rgen},
  institution={IDSIA},
  booktitle={Proceedings of the 23rd international conference on Machine learning},
  pages={369--376},
  year={2006},
  organization={ACM},
  href={https://www.cs.toronto.edu/~graves/icml_2006.pdf}
}

@inproceedings{chorowski2015attention,
  title={Attention-based models for speech recognition},
  author={Chorowski, Jan K and Bahdanau, Dzmitry and Serdyuk, Dmitriy and Cho, Kyunghyun and Bengio, Yoshua},
  booktitle={NIPS},
  pages={577--585},
  year={2015},
  href={https://arxiv.org/abs/1506.07503}
}

@inproceedings{chan2016las,
title = {Listen, Attend and Spell: A Neural Network for Large Vocabulary Conversational Speech Recognition},
author  = {William Chan and Navdeep Jaitly and Quoc V. Le and Oriol Vinyals},
year  = {2016},
pages        = {4960--4964},
booktitle = ICASSP,
href={https://arxiv.org/abs/1508.01211}
}

@inproceedings{hori-etal-2017-joint,
    title = "Joint {CTC}/attention decoding for end-to-end speech recognition",
    author = "Hori, Takaaki  and
      Watanabe, Shinji  and
      Hershey, John",
    editor = "Barzilay, Regina  and
      Kan, Min-Yen",
    booktitle = ACL,
    month = jul,
    year = "2017",
    address = "Vancouver, Canada",
    publisher = "Association for Computational Linguistics",
    url = "https://aclanthology.org/P17-1048",
    doi = "10.18653/v1/P17-1048",
    pages = "518--529"
}

@inproceedings{huang2024lesspeaky,
  title={Less Peaky and More Accurate {CTC} Forced Alignment by Label Priors},
  author={Huang, Ruizhe and Zhang, Xiaohui and Ni, Zhaoheng and Sun, Li and Hira, Moto and Hwang, Jeff and Manohar, Vimal and Pratap, Vineel and Wiesner, Matthew and Watanabe, Shinji and Povey, Daniel and Khudanpur, Sanjeev},
 booktitle=ICASSP,
  pages={11831--11835},
  year={2024}
}

@article{prabhavalkar2023end,
	title={End-to-end speech recognition: A survey},
	author={Prabhavalkar, Rohit and Hori, Takaaki and Sainath, Tara N and Schl{\"u}ter, Ralf and Watanabe, Shinji},
	journal=TASLP,
	volume={32},
	pages={325--351},
	year={2023},
	doi={10.1109/TASLP.2023.3328283},
}

@inproceedings{rousso24_interspeech,
  title     = {Tradition or Innovation: A Comparison of Modern ASR Methods for Forced Alignment},
  author    = {Rotem Rousso and Eyal Cohen and Joseph Keshet and Eleanor Chodroff},
  year      = {2024},
  booktitle = INTERSPEECH,
  pages     = {1525--1529},
  doi       = {10.21437/Interspeech.2024-429},
  issn      = {2958-1796},
}

@InProceedings { schmitt2025:flipped-conformer,
author= {Schmitt, Robin and Zeyer, Albert and Zeineldeen, Mohammad and Schlüter, Ralf and Ney, Hermann},
title= {The Conformer Encoder May Reverse the Time Dimension},
booktitle= ICASSP,
year= 2025,
address= {Hyderabad, India},
month= apr,
note= {Preprint ArXiv:2501.04521 }
}

@misc{microsoft2025phi4mini,
      title={Phi-4-Mini Technical Report: Compact yet Powerful Multimodal Language Models via Mixture-of-LoRAs}, 
      author={Abdelrahman Abouelenin and Atabak Ashfaq and Adam Atkinson and Hany Awadalla and Nguyen Bach and Jianmin Bao and Alon Benhaim and Martin Cai and Vishrav Chaudhary and Congcong Chen and Dong Chen and Dongdong Chen and Junkun Chen and Weizhu Chen and Yen-Chun Chen and Yi-ling Chen and Qi Dai and Xiyang Dai and Ruchao Fan and Mei Gao and Min Gao and Amit Garg and Abhishek Goswami and Junheng Hao and Amr Hendy and Yuxuan Hu and Xin Jin and Mahmoud Khademi and Dongwoo Kim and Young Jin Kim and Gina Lee and Jinyu Li and Yunsheng Li and Chen Liang and Xihui Lin and Zeqi Lin and Mengchen Liu and Yang Liu and Gilsinia Lopez and Chong Luo and Piyush Madan and Vadim Mazalov and Arindam Mitra and Ali Mousavi and Anh Nguyen and Jing Pan and Daniel Perez-Becker and Jacob Platin and Thomas Portet and Kai Qiu and Bo Ren and Liliang Ren and Sambuddha Roy and Ning Shang and Yelong Shen and Saksham Singhal and Subhojit Som and Xia Song and Tetyana Sych and Praneetha Vaddamanu and Shuohang Wang and Yiming Wang and Zhenghao Wang and Haibin Wu and Haoran Xu and Weijian Xu and Yifan Yang and Ziyi Yang and Donghan Yu and Ishmam Zabir and Jianwen Zhang and Li Lyna Zhang and Yunan Zhang and Xiren Zhou},
      year={2025},
      url={https://arxiv.org/abs/2503.01743},
      howpublished={ArXiv 2503.01743}
}

@misc{papi2026doa,
      title={{DOA}: Training-Free Decoder-Only Attention Policy for Long-Form Simultaneous Translation with {SpeechLLMs}},
      author={Sara Papi and Luisa Bentivogli},
      year={2026},
      eprint={2605.31432},
      archivePrefix={arXiv},
      primaryClass={cs.CL},
      url={https://arxiv.org/abs/2605.31432},
      howpublished={ArXiv 2605.31432}
}

@misc{wang2024aeam,
      title={Attention-Constrained Inference for Robust Decoder-Only Text-to-Speech}, 
      author={Hankun Wang and Chenpeng Du and Yiwei Guo and Shuai Wang and Xie Chen and Kai Yu},
      year={2024},
      eprint={2404.19723},
      archivePrefix={arXiv},
      primaryClass={eess.AS},
      url={https://arxiv.org/abs/2404.19723}, 
      howpublished={ArXiv 2404.19723}
}

@misc{mu2026llmforcedaligner,
      title={{LLM-ForcedAligner}: A Non-Autoregressive and Accurate LLM-Based Forced Aligner for Multilingual and Long-Form Speech}, 
      author={Bingshen Mu and Xian Shi and Xiong Wang and Hexin Liu and Jin Xu and Lei Xie},
      year={2026},
      eprint={2601.18220},
      archivePrefix={arXiv},
      primaryClass={cs.SD},
      url={https://arxiv.org/abs/2601.18220}, 
      howpublished={ArXiv 2601.18220}
}

@inproceedings{bain2023whisperx,
  title     = {{WhisperX}: Time-Accurate Speech Transcription of Long-Form Audio},
  author    = {Max Bain and Jaesung Huh and Tengda Han and Andrew Zisserman},
  year      = {2023},
  booktitle = INTERSPEECH,
  pages     = {4489--4493},
  doi       = {10.21437/Interspeech.2023-78},
  issn      = {2958-1796},
}

@InProceedings{radford2023whisper,
  title = 	 {Robust Speech Recognition via Large-Scale Weak Supervision},
  author =       {Radford, Alec and Kim, Jong Wook and Xu, Tao and Brockman, Greg and Mcleavey, Christine and Sutskever, Ilya},
  booktitle = 	 ICML,
  pages = 	 {28492--28518},
  year = 	 {2023},
  editor = 	 {Krause, Andreas and Brunskill, Emma and Cho, Kyunghyun and Engelhardt, Barbara and Sabato, Sivan and Scarlett, Jonathan},
  volume = 	 {202},
  series = 	 {Proceedings of Machine Learning Research},
  month = 	 {23--29 Jul},
  publisher =    {PMLR},
  url = 	 {https://proceedings.mlr.press/v202/radford23a.html},
}

@inproceedings{wagner2024crisperwhisper,
  title     = {{CrisperWhisper}: Accurate Timestamps on Verbatim Speech Transcriptions},
  author    = {Mario Zusag and Laurin Wagner and Bernhad Thallinger},
  year      = {2024},
  booktitle = INTERSPEECH,
  pages     = {1265--1269},
  doi       = {10.21437/Interspeech.2024-731},
  issn      = {2958-1796},
}

@inproceedings{yeh2025whisperaligner,
  title={Whisper Has an Internal Word Aligner},
  author={Yeh, Sung-Lin and Meng, Yen and Tang, Hao},
  booktitle=ASRU,
  year={2025}
}

@inproceedings{ding2019saliency,
    title = "Saliency-driven Word Alignment Interpretation for Neural Machine Translation",
    author = "Ding, Shuoyang  and
      Xu, Hainan  and
      Koehn, Philipp",
    editor = "Bojar, Ond{\v{r}}ej  and
      Chatterjee, Rajen  and
      Federmann, Christian  and
      Fishel, Mark  and
      Graham, Yvette  and
      Haddow, Barry  and
      Huck, Matthias  and
      Yepes, Antonio Jimeno  and
      Koehn, Philipp  and
      Martins, Andr{\'e}  and
      Monz, Christof  and
      Negri, Matteo  and
      N{\'e}v{\'e}ol, Aur{\'e}lie  and
      Neves, Mariana  and
      Post, Matt  and
      Turchi, Marco  and
      Verspoor, Karin",
    booktitle = "Proceedings of the Fourth Conference on Machine Translation (Volume 1: Research Papers)",
    month = aug,
    year = "2019",
    address = "Florence, Italy",
    publisher = "Association for Computational Linguistics",
    url = "https://aclanthology.org/W19-5201/",
    doi = "10.18653/v1/W19-5201",
    pages = "1--12"
}

@inproceedings{mcauliffe2017montreal,
  title     = {{Montreal Forced Aligner}: Trainable Text-Speech Alignment Using {Kaldi}},
  author    = {Michael McAuliffe and Michaela Socolof and Sarah Mihuc and Michael Wagner and Morgan Sonderegger},
  year      = {2017},
  booktitle = INTERSPEECH,
  pages     = {498--502},
  doi       = {10.21437/Interspeech.2017-1386},
  issn      = {2958-1796},
}

@article{pratap2024mms,
author = {Pratap, Vineel and Tjandra, Andros and Shi, Bowen and Tomasello, Paden and Babu, Arun and Kundu, Sayani and Elkahky, Ali and Ni, Zhaoheng and Vyas, Apoorv and Fazel-Zarandi, Maryam and Baevski, Alexei and Adi, Yossi and Zhang, Xiaohui and Hsu, Wei-Ning and Conneau, Alexis and Auli, Michael},
title = {Scaling speech technology to 1,000+ languages},
year = {2024},
publisher = {JMLR.org},
volume = {25},
number = {1},
issn = {1532-4435},
journal = JMLR,
month = jan,
articleno = {97},
numpages = {52}
}

@ARTICLE{rabiner1989tutorial,
  author  = {Lawrence R. Rabiner},
  journal = PROC-IEEE,
  title={A tutorial on hidden {Markov} models and selected applications in speech recognition}, 
  year={1989},
  volume={77},
  number={2},
  pages={257-286},
  doi={10.1109/5.18626}
}

@misc{garofolo1993timit,
  author       = {John S. Garofolo and Lori F. Lamel and William M. Fisher and Jonathan G. Fiscus and David S. Pallett and Nancy L. Dahlgren and Victor Zue},
  title        = {{TIMIT} Acoustic-Phonetic Continuous Speech Corpus},
  howpublished = {Linguistic Data Consortium, Philadelphia, {LDC93S1}},
  year         = {1993},
  doi          = {10.35111/17gk-bn40},
}

@article{pitt2005buckeye,
  title = {The {Buckeye} Corpus of Conversational Speech: Labeling Conventions and a Test of Transcriber Reliability},
  author = {Pitt, Mark A. and Johnson, Keith and Hume, Elizabeth and Kiesling, Scott and Raymond, William},
  year = 2005,
  journal = SPEECHCOMMUNICATION,
  volume = {45},
  number = {1},
  pages = {89--95},
  issn = {0167-6393},
  doi = {10.1016/j.specom.2004.09.001}
}

@article{gales2008application,
author = {Gales, Mark and Young, Steve},
title = {The application of hidden {Markov} models in speech recognition},
year = {2008},
issue_date = {January 2008},
publisher = {Now Publishers Inc.},
address = {Hanover, MA, USA},
volume = {1},
number = {3},
issn = {1932-8346},
doi = {10.1561/2000000004},
journal = {Found. Trends Signal Process.},
month = jan,
pages = {195--304},
numpages = {110}
}

@misc{graves2012transduction,
  author       = {Alex Graves},
  title        = {Sequence Transduction with Recurrent Neural Networks},
  howpublished = {ArXiv:1211.3711, ICML Representation Learning Workshop},
  year         = {2012},
}

@misc{simonyan2014deep,
      title={Deep Inside Convolutional Networks: Visualising Image Classification Models and Saliency Maps}, 
      author={Karen Simonyan and Andrea Vedaldi and Andrew Zisserman},
  howpublished      = {ArXiv:1312.6034; ICLR Workshop Track},
  year      = {2014},
}

@InProceedings{sundararajan2017axiomatic,
  title = 	 {Axiomatic Attribution for Deep Networks},
  author =   {Mukund Sundararajan and Ankur Taly and Qiqi Yan},
  booktitle =ICML,
  pages = 	 {3319--3328},
  year = 	 {2017},
  editor = 	 {Precup, Doina and Teh, Yee Whye},
  volume = 	 {70},
  series = 	 {Proceedings of Machine Learning Research},
  month = 	 aug,
  publisher =    {PMLR},
  url = 	 {https://proceedings.mlr.press/v70/sundararajan17a.html}
}

@InProceedings{shrikumar2017learning,
  title = 	 {Learning Important Features Through Propagating Activation Differences},
  author =       {Avanti Shrikumar and Peyton Greenside and Anshul Kundaje},
  booktitle = ICML,
  pages = 	 {3145--3153},
  year = 	 {2017},
  editor = 	 {Precup, Doina and Teh, Yee Whye},
  volume = 	 {70},
  series = 	 {Proceedings of Machine Learning Research},
  month = 	 aug,
  publisher =    {PMLR},
  url = 	 {https://proceedings.mlr.press/v70/shrikumar17a.html}
}

@inproceedings{ancona2018towards,
title={Towards better understanding of gradient-based attribution methods for Deep Neural Networks},
author={Marco Ancona and Enea Ceolini and Cengiz Öztireli and Markus Gross},
booktitle=ICLR,
year={2018},
url={https://openreview.net/forum?id=Sy21R9JAW}
}

@misc{smilkov2017smoothgrad,
      title={{SmoothGrad}: removing noise by adding noise}, 
      author={Daniel Smilkov and Nikhil Thorat and Been Kim and Fernanda Viégas and Martin Wattenberg},
      year={2017},
  howpublished = {arXiv:1706.03825}
}

@inproceedings{adebayo2018sanity,
 author = {Adebayo, Julius and Gilmer, Justin and Muelly, Michael and Goodfellow, Ian and Hardt, Moritz and Kim, Been},
 booktitle = NEURIPS,
 editor = {S. Bengio and H. Wallach and H. Larochelle and K. Grauman and N. Cesa-Bianchi and R. Garnett},
 pages = {},
 publisher = {Curran Associates, Inc.},
 title = {Sanity Checks for Saliency Maps},
 url = {https://proceedings.neurips.cc/paper_files/paper/2018/file/294a8ed24b1ad22ec2e7efea049b8737-Paper.pdf},
 volume = {31},
 year = {2018}
}

@inproceedings{baevski2020wav2vec,
 author = {Baevski, Alexei and Zhou, Yuhao and Mohamed, Abdelrahman and Auli, Michael},
 booktitle = NEURIPS,
 editor = {H. Larochelle and M. Ranzato and R. Hadsell and M.F. Balcan and H. Lin},
 pages = {12449--12460},
 publisher = {Curran Associates, Inc.},
 title = {{wav2vec 2.0}: A Framework for Self-Supervised Learning of Speech Representations},
 url = {https://proceedings.neurips.cc/paper_files/paper/2020/file/92d1e1eb1cd6f9fba3227870bb6d7f07-Paper.pdf},
 volume = {33},
 year = {2020}
}

@INPROCEEDINGS{shi2021emformer,
  author={Shi, Yangyang and Wang, Yongqiang and Wu, Chunyang and Yeh, Ching-Feng and Chan, Julian and Zhang, Frank and Le, Duc and Seltzer, Mike},
  booktitle=ICASSP, 
  title={Emformer: Efficient Memory Transformer Based Acoustic Model for Low Latency Streaming Speech Recognition}, 
  year={2021},
  volume={},
  number={},
  pages={6783-6787},
  doi={10.1109/ICASSP39728.2021.9414560}
}

@article{erion2021improving,
  title = {Improving Performance of Deep Learning Models with Axiomatic Attribution Priors and Expected Gradients},
  author = {Erion, Gabriel and Janizek, Joseph D. and Sturmfels, Pascal and Lundberg, Scott M. and Lee, Su-In},
  year = 2021,
  month = jul,
  journal = {Nature Machine Intelligence},
  volume = {3},
  number = {7},
  pages = {620--631},
  issn = {2522-5839},
  doi = {10.1038/s42256-021-00343-w}
}

@inproceedings{babu2022xlsr,
  title     = {{XLS-R}: Self-supervised Cross-lingual Speech Representation Learning at Scale},
  author    = {Arun Babu and Changhan Wang and Andros Tjandra and Kushal Lakhotia and Qiantong Xu and Naman Goyal and Kritika Singh and Patrick {von Platen} and Yatharth Saraf and Juan Pino and Alexei Baevski and Alexis Conneau and Michael Auli},
  year      = {2022},
  booktitle = INTERSPEECH,
  pages     = {2278--2282},
  doi       = {10.21437/Interspeech.2022-143},
  issn      = {2958-1796},
}

@misc { phy22-phoneme,
  author       = {Phy, Vitou},
  title        = {Automatic Phoneme Recognition on {TIMIT} Dataset with {Wav2Vec 2.0}},
  year         = 2022,
  publisher    = {Hugging Face},
  version      = {1.0},
  doi          = {10.57967/hf/0125},
  url          = {https://huggingface.co/vitouphy/wav2vec2-xls-r-300m-timit-phoneme}
}

@INPROCEEDINGS{rekesh2023fastconformer,
  author={Rekesh, Dima and Koluguri, Nithin Rao and Kriman, Samuel and Majumdar, Somshubra and Noroozi, Vahid and Huang, He and Hrinchuk, Oleksii and Puvvada, Krishna and Kumar, Ankur and Balam, Jagadeesh and Ginsburg, Boris},
  booktitle=ASRU, 
  title={Fast Conformer With Linearly Scalable Attention For Efficient Speech Recognition}, 
  year={2023},
  volume={},
  number={},
  pages={1-8},
  doi={10.1109/ASRU57964.2023.10389701}
}

@INPROCEEDINGS{Noroozi2024CacheAwareFastConformer,
  author={Noroozi, Vahid and Majumdar, Somshubra and Kumar, Ankur and Balam, Jagadeesh and Ginsburg, Boris},
  booktitle=ICASSP, 
  title={Stateful Conformer with Cache-Based Inference for Streaming Automatic Speech Recognition}, 
  year={2024},
  volume={},
  number={},
  pages={12041-12045},
  doi={10.1109/ICASSP48485.2024.10446861}
}

@InProceedings{xu2023tdt,
  title = 	 {Efficient Sequence Transduction by Jointly Predicting Tokens and Durations},
  author =       {Xu, Hainan and Jia, Fei and Majumdar, Somshubra and Huang, He and Watanabe, Shinji and Ginsburg, Boris},
  booktitle = 	ICML,
  pages = 	 {38462--38484},
  year = 	 {2023},
  editor = 	 {Krause, Andreas and Brunskill, Emma and Cho, Kyunghyun and Engelhardt, Barbara and Sabato, Sivan and Scarlett, Jonathan},
  volume = 	 {202},
  series = 	 {Proceedings of Machine Learning Research},
  month = 	 {23--29 Jul},
  publisher =    {PMLR},
  url = 	 {https://proceedings.mlr.press/v202/xu23g.html}
}

@inproceedings{zhang2023speechgpt,
    title = "{S}peech{GPT}: Empowering Large Language Models with Intrinsic Cross-Modal Conversational Abilities",
    author = "Zhang, Dong  and
      Li, Shimin  and
      Zhang, Xin  and
      Zhan, Jun  and
      Wang, Pengyu  and
      Zhou, Yaqian  and
      Qiu, Xipeng",
    editor = "Bouamor, Houda  and
      Pino, Juan  and
      Bali, Kalika",
    booktitle = "Findings of the Association for Computational Linguistics: EMNLP 2023",
    month = dec,
    year = "2023",
    address = "Singapore",
    publisher = "Association for Computational Linguistics",
    url = "https://aclanthology.org/2023.findings-emnlp.1055/",
    doi = "10.18653/v1/2023.findings-emnlp.1055",
    pages = "15757--15773"
}

@misc{chu2023qwenaudio,
  title={{Qwen-Audio}: Advancing Universal Audio Understanding via Unified Large-Scale Audio-Language Models},
  author={Chu, Yunfei and Xu, Jin and Zhou, Xiaohuan and Yang, Qian and Zhang, Shiliang and Yan, Zhijie  and Zhou, Chang and Zhou, Jingren},
  howpublished={arXiv:2311.07919},
  year={2023}
}

@inproceedings{peng2024owsmctc,
    title = "{OWSM}-{CTC}: An Open Encoder-Only Speech Foundation Model for Speech Recognition, Translation, and Language Identification",
    author = "Peng, Yifan  and
      Sudo, Yui  and
      Shakeel, Muhammad  and
      Watanabe, Shinji",
    booktitle = ACL,
    year = "2024",
    month= {8},
    pages = "10192--10209",
    url = "https://aclanthology.org/2024.acl-long.549",
    doi = "10.18653/v1/2024.acl-long.549"
}

@inproceedings{peng2025owsmv4,
  title={{OWSM} v4: Improving Open Whisper-Style Speech Models via Data Scaling and Cleaning},
  author={Yifan Peng and Shakeel Muhammad and Yui Sudo and William Chen and Jinchuan Tian and Chyi-Jiunn Lin and Shinji Watanabe},
  booktitle=INTERSPEECH,
  year={2025},
}

@inproceedings{puvvada2024canary,
  title     = {Less is More: Accurate Speech Recognition {\&} Translation without Web-Scale Data},
  author    = {Krishna C. Puvvada and Piotr Żelasko and He Huang and Oleksii Hrinchuk and Nithin Rao Koluguri and Kunal Dhawan and Somshubra Majumdar and Elena Rastorgueva and Zhehuai Chen and Vitaly Lavrukhin and Jagadeesh Balam and Boris Ginsburg},
  year      = {2024},
  booktitle = INTERSPEECH,
  pages     = {3964--3968},
  doi       = {10.21437/Interspeech.2024-2294},
  issn      = {2958-1796},
}

@inproceedings{chen2025owls,
title={{OWLS}: Scaling Laws for Multilingual Speech Recognition and Translation Models},
author={William Chen and Jinchuan Tian and Yifan Peng and Brian Yan and Chao-Han Huck Yang and Shinji Watanabe},
booktitle=ICML,
year={2025},
url={https://openreview.net/forum?id=xnPW7yYomF}
}

@misc{mistral2025voxtral,
  author       = {{Mistral AI}},
  title        = {{Voxtral}},
  howpublished = {arXiv:2507.13264},
  year         = {2025},
}

@INPROCEEDINGS{yang2022torchaudio,
  author={Yang, Yao-Yuan and Hira, Moto and Ni, Zhaoheng and Astafurov, Artyom and Chen, Caroline and Puhrsch, Christian and Pollack, David and Genzel, Dmitriy and Greenberg, Donny and Yang, Edward Z. and Lian, Jason and Hwang, Jeff and Chen, Ji and Goldsborough, Peter and Narenthiran, Sean and Watanabe, Shinji and Chintala, Soumith and Quenneville-Bélair, Vincent},
  booktitle=ICASSP, 
  title={{TorchAudio}: Building Blocks for Audio and Speech Processing}, 
  year={2022},
  volume={},
  number={},
  pages={6982-6986},
  doi={10.1109/ICASSP43922.2022.9747236}
}

@INPROCEEDINGS{hwang2023torchaudio,
  author={Hwang, Jeff and Hira, Moto and Chen, Caroline and Zhang, Xiaohui and Ni, Zhaoheng and Sun, Guangzhi and Ma, Pingchuan and Huang, Ruizhe and Pratap, Vineel and Zhang, Yuekai and Kumar, Anurag and Yu, Chin-Yun and Zhu, Chuang and Liu, Chunxi and Kahn, Jacob and Ravanelli, Mirco and Sun, Peng and Watanabe, Shinji and Shi, Yangyang and Tao, Yumeng},
  booktitle=ASRU, 
  title={{TorchAudio 2.1}: Advancing Speech Recognition, Self-Supervised Learning, and Audio Processing Components for {PyTorch}},
  year={2023},
  volume={},
  number={},
  pages={1-9},
  doi={10.1109/ASRU57964.2023.10389648}
}

@InProceedings { zeyer2018:asr-attention,
author= {Zeyer, Albert and Irie, Kazuki and Schlüter, Ralf and Ney, Hermann},
title= {Improved training of end-to-end attention models for speech recognition},
booktitle= {Interspeech},
year= 2018,
address= {Hyderabad, India},
month= sep,
booktitlelink= {http://interspeech2018.org/}
}

@Misc { schmitt2026:speech-llm,
title= {{LLMs} and Speech: Integration vs. Combination},
author= {Schmitt, Robin and Zeyer, Albert and Zeineldeen, Mohammad and Schlüter, Ralf and Ney, Hermann},
howpublished= {Arxiv:2603.15045},
month= mar,
year= 2026,
url = {https://arxiv.org/abs/2603.15045}
}

\end{document}